\documentclass[journal ]{new-aiaa}
\usepackage[utf8]{inputenc}
\usepackage{textcomp}
\usepackage{graphicx}
\usepackage[version=4]{mhchem}
\usepackage{siunitx}
\usepackage{longtable,tabularx}
\usepackage{tikz}
\usetikzlibrary{arrows.meta, positioning}
\usepackage{algorithm}
\usepackage{algpseudocode}
\usepackage{hyperref}
\usepackage{array}
\AtBeginEnvironment{algorithm}{%
  \singlespacing
}

\newcommand {\R}{\mathbb{R}}
\newcommand {\Psp}{\mathbb{P}}
\newcommand {\Esp}{\mathbb{E}}
\newcommand {\Nsp}{\mathbb{N}}

\usepackage{color}

\title{Path Integral Particle Filtering for Hybrid Systems via Saltation Matrices}
\author{Karthik Shaji \footnote{Graduate Student, Daniel Guggenheim School of Aerospace Engineering}, Sreeranj Jayadevan \footnote{MathWorks}, Bo Yuan \footnote{Graduate Student, Daniel Guggenheim School of Aerospace Engineering}, Hongzhe Yu \footnote{PlusAI}, Yongxin Chen \footnote{Associate Professor, Daniel Guggenheim School of Aerospace Engineering}}
\affil{Georgia Institute of Technology, Atlanta, GA, 30332}

\begin{document}
\maketitle

\begin{abstract}
State estimation for hybrid systems that undergo intermittent contact with their environments, such as extraplanetary robots and satellites undergoing docking operations, is difficult due to the discrete uncertainty propagation during contact. 
To handle this propagation, this paper presents an optimal-control-based particle filtering method that leverages saltation matrices to map out uncertainty propagation during contact events. 
By exploiting a path integral filtering framework that exploits the duality between smoothing and optimal control, the resulting state estimation algorithm is robust to outlier effects, flexible to non-Gaussian noise distributions, and handles challenging contact dynamics in hybrid systems. 
To evaluate the validity and consistency of the proposed approach, this paper tests it against strong baselines on the stochastic dynamics generated by a bouncing ball and spring loaded inverted pendulum.
% This paper proves the validity and consistency of the proposed approach through simulations on the stochastic dynamics generated by a bouncing ball and spring loaded inverted pendulum, from which connections are drawn to Aerospace systems.
% This work offers a computationally efficient and reliable estimation for hybrid systems with stochastic dynamics.
% The robus
% The path integral particle filter is computationally efficient, and saltation matrices yield more accurate results for uncertainty propagation across transition events in hybrid systems. 
% This work offers a computationally efficient and reliable estimation algorithm for hybrid systems with stochastic dynamics. We also present extensive experimental results demonstrating that our approach consistently outperforms strong baselines across multiple settings.

\end{abstract}
% I checked with other papers in JGCD, and most don't have nomenclatures

\section{Introduction}
% {\hongzhe{ 1. What's the problem we consider? Why is it important? 2. What are the existing methods? What are their limitations? 3. What's our goal? What's our method? What are the potential advantages of our algorithms? 4. Literature review. What's the relation between our method and existing methods?}}
The objective of filtering in dynamic systems is to estimate the ``true'' state of a given system in the presence of sensor measurements. In continuous-time Aerospace systems, three commonly used paradigms are the Kalman filter (KF) \cite{kalman1960}, the Extended Kalman filter (EKF) \cite{julier1997}, as well as the Unscented Kalman filter (UKF) \cite{julier1995, ekf_ukf_spacecraft}. One common downside to all of these algorithms is that they intrinsically assume the noise distribution is close to Gaussian, and their performance degrades under non-Gaussian noise \cite{Qi2023MaximumCorrentropyEKF, ZHAO2022110410}. For systems that exhibit colliding contact with their environments, such as bipedal and quadrupedal robots \cite{zhao_ames_16}, robotic manipulators \cite{Koval2015MPF}, and impact-driven mechanical systems \cite{Payne-2024}, this difficulty is only further amplified. These hybrid systems transition between modes based on contact events, making state estimation particularly challenging due to sudden, discontinuous changes in dynamics. This area of work has shown increasing value, with application to extraplanetary robotics, such as the super-ball bot and tensegrity robotics projects \cite{SunSpiral2015SuperBallBot, Garanger2021SoftTensegrity}, trajectory optimization for planetary landing schemes (which make contact on landing) \cite{Saranathan2018RASHS}, and Rendezvous, Proximity Operations and Docking (RPOD) \cite{soderlund_rpod}.

\begin{figure}[H]
\centering

\begin{tikzpicture}[>=Stealth, node distance=4cm]

\tikzset{
  mode/.style = {circle, draw, minimum width=2.5cm, align=center}
}

% Nodes
\node[mode] (q0) {$q_0$\\$\dot{x} = f_0(x)$\\$x \in Inv(q_0)$};
\node[mode, right=of q0] (q1) {$q_1$\\$\dot{x} = f_1(x)$\\$x \in Inv(q_1)$};

% Initial arrow
\draw[->] (q0.north) ++(0,1)
    node[above, align=center] {$(q_0,x_0) \in Init$}
    -- (q0.north);

% Transitions
\draw[->, bend left=20]
    (q0) to
    node[above, align=center] {$G(q_0,q_1)$\\$x^{+} = R(q_0,q_1,x)$}
    (q1);

\draw[->, bend left=20]
    (q1) to
    node[below, align=center] {$G(q_1,q_0)$\\$x^{+} = R(q_1,q_0,x)$}
    (q0);

\end{tikzpicture}
\caption{Hybrid automaton with two modes}
\label{hybrid_automata}
\end{figure}
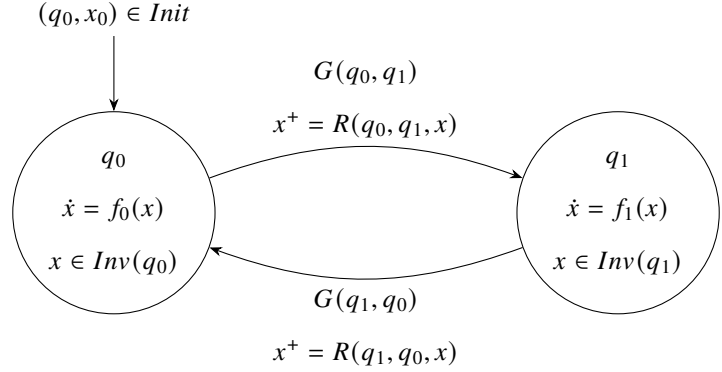

One method to get around challenges with non-Gaussian noise distributions is through particle filtering \cite{gordon1993}, also known as Monte Carlo filtering, where a given state is estimated by an ensemble of particles, with their own weight. One commonly used approach is the Sequential Importance Resampling (SIR) filter, where the particle weight update is performed through approximating inter-particle relative posterior probabilities \cite{kitagawa1996}. It is known that as the number of particles increases, the state estimate more closely resembles the posterior distribution of the noise \cite{delmoral1997}. One major problem with particle filters is the so-called ``weight degeneracy problem,'' where individual particles can possess a high relative weight compared to the rest, and the estimation characteristics end up degrading \cite{murata_particle_degeneracy}. One approach to mitigating weight degeneracy is through increasing particle counts, either in a fixed or adaptive manner \cite{Fox2001KLDSamplingAP}, as it increases the time it takes for weight degeneracy to occur, but that leads to higher computational costs \cite{Elvira2015AdaptingTN}. Sequential Importance Resampling aims to get around this issue by periodically resampling the particle weights and states once a resampling threshold is reached, but
% this operation is rather expensive \cite{Douc_resampling_comp}. Furthermore, 
this operation does not scale well to high-dimensional problems, where higher-order state dimensions and greater particle counts are needed \cite{ObstaclestoHighDimensionalParticleFiltering}.\\
To address challenges with weight degeneracy, we follow the path integral particle filtering framework and leverage the duality between estimation and optimal control for continuous-time systems \cite{kim2020optimal}. A useful lens for understanding nonlinear smoothing is its duality with stochastic optimal control: rather than treating inference purely as Bayesian filtering, state estimation can be posed as an optimal control / variational problem over state trajectories \cite{kim2020optimal}. Optimal feedback laws and their associated posterior path distributions can be approximated using Monte-Carlo sampling, thereby connecting smoothing/inference with sampling-based control computation.\\
Using this duality, we compute a posterior estimate by computing a suboptimal solution to an optimal control problem, developing a particle filter with better performance. The particular advantage of our work is the consideration of the time history of state evolution \cite{particleFilteringBasics}, whereas most commonly used particle filtering approaches only use the present sampling \cite{kitagawa1996}. The major advantage of this approach is that it enables the correction of previous outliers and allows the algorithm to mitigate particle degeneracy and reduce the occurrence of expensive resampling steps. \\
We build upon \cite{ZHANG2023110894}, which proposed a path integral particle filtering formulation for continuous-time systems. We extend this prior work to hybrid systems, characterized by continuous and discrete actions \cite{allenback1993hybrid}, in contact-driven domains. Traditional classical control and estimation methods struggle with this due to their underlying smooth and continuous assumptions \cite{bledt2018}. With state estimation, one has to not only consider the present mode but also avoid potential combinatorial problems from multiple mode mismatches \cite{zhang2020} between nominal and estimated dynamics in propagation calculations. In contrast to systems described solely by linear or nonlinear continuous-time dynamics, hybrid systems include discrete modes that must be specified alongside the continuous equations to fully characterize the system. Hybrid systems are represented using hybrid automata \cite{HybridAutomata}, with each mode defined by its own dynamics, as shown in Fig. \ref{hybrid_automata}. Mode transitions are triggered by guard conditions, may apply reset maps, and, for linear systems, involve switching between different sets of system matrices. For nonlinear systems, this involves changing the stochastic differential equations associated with each mode \cite{Hybridsystems}.\\
Existing estimation methods, including the previous continuous-time path integral particle filter \cite{ZHANG2023110894},  depend on understanding how probability distributions or uncertainties evolve through system dynamics \cite{appliedStochasticCh3}. This task becomes significantly more complicated in hybrid systems, where mode transitions may change the dimensionality of the state or modify the measurable states \cite{HybridEstimation}. State estimation for such systems has been difficult, with novel methods being employed for many scenarios \cite{ContextBasedStateEstimation,HybridMarkov}. Hwang et al. (2005) presented the Multiple Model Adaptive Estimation (MMAE) method \cite{Hamsa}, in which Kalman filters operate in parallel with a mode-probability mixing mechanism. They modeled the aircraft as a discrete-time stochastic linear hybrid system, where multiple modes represent various flight regimes. While MMAE shows promising performance, it shares the same fundamental limitation as the previously discussed approaches: it is restricted to linear system models with Gaussian noise. More recently \cite{leine2013} has also leveraged the concept of saltation matrix, a tool for analysis of non-smooth systems, in the context of Kalman-like Filters, developing the Salted Kalman Filter (SKF) \cite{kong2021}. One key advantage of using the saltation matrix is that it gives a first-order approximation of the uncertainty propagation and dynamics changes from hybrid transitions, purely by using the Jacobian of the reset map \cite{KongKalman}. By pairing that with a new dynamics model for the new mode, one can properly estimate a Gaussian noise distribution about a contact event.
For the optimal control problem, the framework developed in \cite{kong2023saltation} was used for a salted iLQR algorithm, but with a different cost function. The path integral control for hybrid systems from \cite{yu2024path} is also employed as a basis for our sampling-based particle control method. The goal here is to provide an algorithm that can perform accurate state estimation for hybrid dynamical systems, even when the noise follows non-Gaussian distributions, addressing a drawback in the method presented by \cite{KongKalman}. They avoided particle filters because of their computational cost, a concern that \cite{Chen2022} later improved upon. In hybrid systems, the state dimension may change across modes, making state estimation especially challenging at transition points. Hence, as mentioned before, the reference-trajectory extension approach introduced in \cite{yu2024path,kong2021} is followed, which provides a consistent way to extend trajectories across modes and improves estimation accuracy. \\
The results from the algorithm, the Salted Path Integral Particle Filter (SPIPF) are more accurate and computationally efficient compared to a traditional particle filter algorithm such as a multi-mode version of the SIR filter \cite{kitagawa1996}. The results were compared against the baselines and the true posterior results. This was tested on two different hybrid systems: the bouncing ball system and the Spring Loaded Inverted Pendulum (SLIP) system. The first system has linear dynamics, whereas the SLIP system has nonlinear dynamics \cite{kong2021}. The primary goal of our paper is to demonstrate improved contact estimation compared to other particle filter baselines on Gaussian noise distributions, whilst relying on the distribution-agnosticity of particle filters to generalize performance under non-Gaussian noise. Thus, the contributions of this paper are twofold. \\
Our paper is organized as follows: Section II discusses the mathematics and relevant notations, Section III discusses our new problem formulation, Section IV discusses the algorithm and its implementation, Section V explains the experiments used to validate performance, Section VI provides results compared to relevant baselines, and Section VII provides our conclusion.

\section{Notation and Preliminaries}
\subsection{Smooth Path Integral Particle Filtering}
The goal of this section of the paper is to discuss how a smooth filtering problem can be converted into an analogous optimal control problem. In \cite{ZHANG2023110894}, the authors used this duality to illustrate how controlled processes can be used to convert a posterior estimation problem into an optimal control one. In our set of stochastic dynamics, $X_t$ is the nominal stochastic model, and $Y_t$ is the measurement. $X_t$ has a stochasticity described with the Wiener process $W_t \in \R^p$ where the initial state $X_0$ follows the prior distribution $\nu_0$, while the measurement $Y_t \in \R^p$ is corrupted by the standard Wiener process $B_t \in \R^p$ weighted by $\sigma_B > 0$. For brevity, $h(t, X_t)$ is denoted by $h_t$, and the two Wiener processes $W_t$ and $B_t$ are defined as independent. The goal of the presented smoothing problem is a sequential filtering problem aiming to find the posterior distribution of $X_t$ given the history of $Y_t$ for $0 \leq t \leq T$, parameterized by the following,
\begin{subequations}
    \begin{equation}\label{smooth_stoch_dyn}
        dX_t = b(t, X_t)dt + \sigma(t, X_t)\sqrt{\epsilon} dW_t, \quad X_0 \sim \nu_0
    \end{equation}
    \begin{equation}\label{smooth_stoch_meas}
        dY_t = h(t, X_t)dt + \sigma_BdB_t, \quad Y_0 = 0.
    \end{equation}
\end{subequations}
The authors in \cite{kim2020optimal} found that this type of problem could be reformulated as a stochastic optimal control problem. Let us denote the measure over the path space $\Omega$ for the nominal dynamics in Eq. \eqref{smooth_stoch_dyn} induced by $\mathcal{P}$, serving as the prior for a Bayesian inference problem, while the posterior distribution is denoted by $\mathcal{Q}^Y$.
By the pathwise Kallianpur-Striebel formula \cite{handel2006}, the ratio between $\mathcal{Q}^Y$ and $\mathcal{P}$ enjoys an explicit form as
\begin{equation}\label{smooth_ks_bayes}
    \frac{d\mathcal{Q}^Y}{d\mathcal{P}} \propto \exp{\biggl\{-\frac{1}{\sigma_B^2}\left[\int_0^TY_tdh_t + \frac{1}{2} \lVert h_t \rVert^2 dt - Y_Th(T, X_T)\right]\biggl\}}.
\end{equation}
From Equation~\eqref{smooth_ks_bayes}, the right-hand side can be interpreted as the likelihood of the observation process given the signal path. Let the path distribution of the controlled process be denoted by
\begin{equation}\label{P_tilde_param}
    d\tilde{X}_t = b(t, \tilde{X_t})dt + \sigma(t, \tilde{X_t})(u_tdt + \sqrt{\epsilon}dW_t), \quad \tilde{X_0} \sim \pi_0
\end{equation}
as $\tilde{\mathcal{P}}$. As derived in \cite{KimSmooth}, Equation~\eqref{smooth_ks_bayes} can be rewritten as the variational form of a smoothing problem that seeks the distribution $\tilde{\mathcal{P}}$ minimizing the following variational objective:
\begin{equation}\label{variational_inf}
   \Esp_{\tilde{\mathcal{P}}}\biggl(\log{\frac{d\tilde{\mathcal{P}}}{d\mathcal{Q}^Y}}\biggl) =
    \Esp_{\tilde{\mathcal{P}}}\biggl(\log\frac{d\tilde{\mathcal{P}}}{d\mathcal{P}} - \log{\frac{d\mathcal{Q}^Y}{d\mathcal{P}}}\biggl) = 
    {\rm KL}(\tilde{\mathcal{P}} \parallel \mathcal{P}) - \Esp_{\tilde{\mathcal{P}}}\biggl(\log{\frac{d\mathcal{Q}^Y}{d\mathcal{P}}}\biggl),
\end{equation}
where the KL divergence is defined between the path space of the controlled and uncontrolled probability distributions $\tilde{\mathcal{P}}$ and $\mathcal{P}$
\begin{equation}\label{KL_def}
         {\rm KL}(\tilde{\mathcal{P}} || \mathcal{P}) = \int d\tilde{\mathcal{P}} \log{\frac{d\tilde{\mathcal{P}}}{d\mathcal{P}}}.
\end{equation}
Girsanov theorem \cite{Simo08} can now be applied, but to the filtering problem, with the key difference that Equation~\eqref{P_tilde_param} does not follow the same distribution as Equation~\eqref{smooth_stoch_dyn}, from which we arrive at 
\begin{equation}\label{smooth_girsanov_filtering}
    {\rm KL}(\tilde{\mathcal{P}} \parallel \mathcal{P}) = \Esp\biggl(\int_0^T \frac{1}{2 \epsilon}\lvert \lvert u_t \rvert \rvert^2 dt\biggl) + {\rm KL}(\pi_0 \parallel \nu_0).
\end{equation}
Equation~\eqref{smooth_ks_bayes} and Equation~\eqref{smooth_girsanov_filtering} can now be plugged into Equation~\eqref{variational_inf}, arriving at our optimal control formulation \cite{kim2020optimal},
\begin{equation}\label{smooth_oc_analog}
    \min\limits_{u, \pi_0}\Esp\biggl(\int_0^T l(t, X_t, u_t) + \Psi_T(X_t)\biggl) + {\rm KL}(\pi_0 \parallel \nu_0),
\end{equation}
where the extra term ${\rm KL}(\pi_0 \parallel \nu_0)$ comes from the difference in initial distributions, and $l(t, X_t, u_t)$, is defined as  
\begin{equation}\label{control_running_cost_simp}
    l(t, X_t, u_t) \triangleq g(t, X_t) + \frac{1}{2 \epsilon } \lvert \lvert u_t \rvert \rvert ^2 dt,
\end{equation}
where  $g(t, X_t)$, $\Psi_T(x)$ are
\begin{equation}\label{smooth_oc_filtering_sub}
    \begin{aligned}
        g(t, X_t) = \frac{1}{2 \sigma_B^2}\lvert \lvert h(t, X_t)\rvert \rvert^2dt + \frac{1}{\sigma_B^2}Y_tdh(t, X_t), \\
        \Psi_T(x) = -\frac{1}{\sigma_B^2}Y_Th(T, X_T).
    \end{aligned}
\end{equation}

\subsection{Hybrid Systems and the Linearization of Jump Dynamics}
\subsubsection{Hybrid Dynamical Systems}
A Hybrid system, per \cite{Grossman1993Hybrid}, \cite{Posa2014Direct}, is defined by the tuple: $\mathcal{H} := \{\mathcal{I}, \mathcal{D}, \mathcal{F}, \mathcal{G}, \mathcal{R}\}$, where:
\begin{itemize}
    \item $\mathcal{I} := \{I_1, I_2,..., I_{N_I}\} \subset \Nsp$ is a finite set of modes. 
    \item $\Gamma \subset \mathcal{J} \times \mathcal{J}$ is the set of discrete transitions which form a directed graph structure over $\mathcal{J}$
    \item $\mathcal{D}$ is the set of continuous domains which represent the state spaces, with pairings of nodes to modes $(D_j, I_j)$.
    \item $\mathcal{F}$ is the set of flows which define the smooth dynamics $F_j$ in each mode $I_j$
    \item $\mathcal{G} := \Pi_{(I, J) \, \in \, \Gamma} G_{(I, J)}(t)$ is the collection of guards where a specific $G_{(I, J)}(t) \subset D_I$ for each $(I, J) \in \Gamma$ is defined as a sublevel set of a $C^r$ function.
    \item $\mathcal{R}$ is the set of reset functions $R_{jk}$ mapping the state from $D_j$ to $D_k$ when the guard condition $G_{jk}$ is met.
\end{itemize}
A transition from mode $I_j$ to mode $I_k$ occurs at state $X_t^- := X(t^-)$ when the guard condition $g_{jk}$
\begin{equation}\label{hybrid_guard}
    g_{jk}(t^-, X_t^-) \leq 0, X_t^- \in I_j.
\end{equation}
In turn, the jump dynamics are defined as
\begin{equation}\label{reset_dyn}
    X_t^+ = R_{jk}(X_t^-) \in D_k,
\end{equation}
where $R_{jk} \in \mathcal{R}$ is the corresponding reset function which maps between the two continuous domains when the guard condition $g_{jk}$ is met. The notation $(t^-, X_t^-)$ and $(t^+, X_t^+)$ is used to represent the pre-impact and post-impact time-state pairs.
\subsubsection{Stochastic Optimal Control with Hybrid Transitions}
Let us define a controlled smooth flow in mode $I_j$ as
\begin{equation}
    dX_t^j = F_j(t, X_t^j)dt + \sigma_j(t, X_t^j)(u^j(t, X_t^j)dt + \sqrt{\epsilon}dW_t^j).
\end{equation}
The corresponding uncontrolled flow is defined, similar to Equation~\eqref{smooth_stoch_dyn}, in this mode $I_j$ as
\begin{equation}
    dX_t^j = F_j(t, X_t^j)dt + \sigma_j(t, X_t^j)\sqrt{\epsilon}dW_t^j.
\end{equation}
Now, denote the period of $[t_j^+, t_{j + 1}^-]$ to be in mode $j$, and denote the start and end period of this sequence as $[0, T]$. In other words
\begin{equation}
    \begin{aligned}
        &X_t^j \in I_j, \forall t \in [t_j^+, t_{j + 1}^-], 
        \\
        &X^{j + 1}(t^+_{j + 1}) = R_{j, j + 1}(X^j(t_j^-)).
    \end{aligned}
\end{equation}
Let us consider the stochastic optimal control problem over a series of different modes, as in \cite{yu2024path}. Here, the order of the hybrid transitions is identified $N_j$ prior to the simulation, causing its cost function to become
\begin{equation}\label{hybrid_oc}
            \min\limits_{u}\mathcal{J} \triangleq \Esp\left[\sum_{j = 0}^{N_{J - 1}} \int_{t_j^+}^{t_{j + 1}^-}l(t, X_t^j, u_t^j) + \Psi_T\right],
\end{equation}
Now, let $F^H_{\Delta, i}(X_i^j, u_i)$ represent the hybrid transition for $x$ from $t_i$ to $t_{i + 1}$. If the guard function $g_{j, j + 1}(t_{i + 1}, X_{t_{i + 1}}) > 0$, then $X^j_{t_{i + 1}} \in I_j$ is still in $I_j$, therefore
\begin{equation}\label{curr_mode_jump}
        X^j_{t_{i + 1}} = F^H_{\Delta, i}(X_{t_i}^j, u_{t_i}^j) \approx X_{t_i}^j + (F_j(t_i, X^j_{t_i}) + \sigma_j(X_{t_i}^j)u_{t_i}^j))\Delta t + \sigma_j(X_{t_i}^j)\Delta W_{t_i}^j.
\end{equation}
On the other hand, if there is a jump at some $t_j^- \in [t_i, t_{i + 1}]$, then $t_j^-$ is approximated as $t_{i + 1}$, which holds up for substantially small timesteps. The hybrid dynamics with a jump are
\begin{equation}\label{next_mode_jump}
        X_{t_{i + 1}}^{j + 1} = F_{\Delta, i}^H(X_{t_i}^j, u_{t_i}^j) 
        \approx R_{j, j + 1}(X_{t_i}^j + F_j(t_i, X_{t_i}^j) + \sigma_j(X_{t_i}^j) u_{t_i}^j)\Delta t +
        \sigma_j(X_{t}^{j})\Delta W_{t_i}^{j}).
\end{equation}
\section{Problem Formulation}
The path integral particle filter formulation \cite{ZHANG2023110894} is modified so that it can be applied for the purpose of hybrid filtering, for which we leverage the principles from Equation~\eqref{hybrid_oc}. 
We will discuss hybrid smoothing and then extend the path integral formulation done in \cite{ZHANG2023110894} and \cite{yu2024path} to cover this area through hybrid optimal control. For notation purposes, $j$ is used frequently, which denotes "for specific-mode" in later contexts.

\subsection{Smoothing for Hybrid Dynamical Systems}
First, a standard diffusion process is considered in the mode $\mathcal{I}_j$. Similar to Equation~\eqref{smooth_stoch_dyn} and Equation~\eqref{smooth_stoch_meas}, the measurement $Y_t \in \R ^p$ is corrupted by the standard Wiener process $B_t \in \R^p$
\begin{subequations}
    \begin{equation}\label{hybrid_stoch_dyn}
        dX_t^j = F^j(t, X_t^j)dt + \sigma^j(t, X_t^j)\sqrt{\epsilon}dW_t^j, \quad X^j_0 \sim \nu^j_0
    \end{equation}
    \begin{equation}\label{hybrid_stoch_meas}
        dY_t^j = H^j(t, X_t^j)dt + \sigma_{B}^j dB_t, \quad Y^j_{0} = 0.
    \end{equation}
\end{subequations}
A transition from mode $I_j$ to mode $I_k$ happens in the dynamic state $X_t^- := X(t^-)$ as in Equation~\eqref{hybrid_guard_dyn}
\begin{equation}\label{hybrid_guard_dyn}
    g_{jk}(t^-, X_t^-) \leq 0, X_t^- \in I_j.
\end{equation}
Although the measurement $Y_t$ is a nonlinear function of $X_t$, it is assumed that it shares the same mode, as would be consistent with an observation of a state. The jump dynamics are defined in the same way as Equation~\eqref{reset_dyn}. Here, no assumption is made that all modes follow the same noise distribution. Furthermore, there is the problem of potential mode mismatch between the nominal dynamics of a roll out and the computed posterior prediction, which will be discussed later. Each mode is assumed to be independent of the others prior to transition (resulting in independent, disjoint time intervals), which allows the use of the Pathwise Kallianpur–Striebel formula. For each mode, the process evolves in a right-continuous space with left limits everywhere, consistent with the definition of a càdlàg process. Therefore, it can be used summatively across all the modes, arriving at
\begin{equation}\label{KS_summative}
    \tilde{Z}_T\left(X_{[t_j^+, t_{j + 1}^-]}, y\right) = \text{exp}\biggl(\sum_{j=0}^{N_j}\biggl(H^j(X_T) \cdot Y_T^j - \int_{t_j^+}^{t_{j + 1}^-} Y_t^j \cdot dH^j(X_t) - \frac{1}{2} \int_{t_j^+}^{t_{j + 1}^-} \lvert \lvert H^j(X_t)\rvert|^2 dt\biggl)\biggl).
\end{equation}
As done previously, $\frac{d\mathcal{Q}^Y}{d\mathcal{P}}$ is equivalent to $\tilde{Z}_T\left(X_{[0, T]}, y\right)$. Here, the following substitution can be made, for the purposes of simplifying notation
\begin{equation}\label{Y_T_sub}
    \tilde{Y}_t = \frac{Y_t^j}{\sigma^j_{B}}.
\end{equation} 
From there, substituting and plugging into Equation~\eqref{KS_summative}, the following equation is derived
\begin{equation}\label{ks_posterior_hybrid}
        \frac{d\mathcal{Q}^Y}{d\mathcal{P}} \propto \exp\biggl(\sum_{j=0}^{N_j}\biggl\{\biggl[\int_{t_j^+}^{t_{j + 1}^-}\mathcal{L}^j(t, X_t, \tilde{Y}_t)\biggl]\biggl\}\biggl).
\end{equation}
For future reference, the interior of the summation is defined as the following "filtering cost per mode" $\mathcal{L}_j$
\begin{equation}\label{general_cost}
    \mathcal{L}^j(t, X_t, Y_t) \triangleq -\frac{1}{\sigma_{B}^{j, 2}}\left(Y_t^jdH^j +
        \frac{1}{2}\lvert\lvert H^j \rvert \rvert^2 dt - Y_T^jH^j(T, X_T)\right),
\end{equation}
for the purpose of simplifying further analysis.
\subsection{Solving Hybrid Filtering Through Hybrid Optimal Control}
The application of Equation~\eqref{hybrid_oc} to the filtering problem in Equation~\eqref{ks_posterior_hybrid} is now discussed. The key difference between this and the previous result for the control problem is that $\mathcal{P}_u$ and $\mathcal{P}_0$ share the same initial distribution while $\tilde{X}_0$ and $X_0$ do not. Hence, the additional KL term denotes the differences in those initial distributions. Here, the result of Equation~\eqref{variational_inf} and Equation~\eqref{smooth_girsanov_filtering} can be applied into the context of each individual hybrid mode. Substituting those into Equation~\eqref{ks_posterior_hybrid}, yields the following variational equation, for each mode
\begin{equation}\label{variational_hybrid}
        \Esp_{\tilde{P}}\biggl\{\log{\frac{d\tilde{\mathcal{P}}}{d\mathcal{Q}^Y}}\biggl\} = \Esp_{\tilde{\mathcal{P}}}\biggl\{\int_{t_j^+}^{t_{j + 1}^-} \frac{1}{2 \epsilon}\lvert \lvert u_t \rvert \rvert^2 dt\biggl\} + KL(\pi^j_{0} \parallel \nu^j_{0})
        -\Esp_{\tilde{\mathcal{P}}}\biggl\{ \log\biggl(\exp\biggl\{\int_{t_j^+}^{t_{j + 1}^-}\mathcal{L}^j(t, X_t, Y_t)\biggl\}\biggl)\biggl\}.
\end{equation}
Once this expression is expanded, it is placed into the summation and turned into the equivalent minimization problem over the control input and initial distribution $\pi_0$. The key note here though is that the problem involving minimizing the expectation from this initial distribution, the initial distribution is dependent on each state $j$. In other words, this can be thought of as the problem of minimizing $J$ independent noise distributions over each mode. This yields
\begin{equation}\label{hybrid_min_xt}
    \min_{u, \pi_0} \Esp \sum_{j=0}^{N_j}\biggl\{\int_{t_j^+}^{t_{j + 1}^-}\frac{1}{2 \epsilon }\lvert \lvert u\rvert \rvert^2dt - \mathcal{L}^j(t, X_t, Y_t) + KL(\pi^j_{0} \parallel  \nu^j_{0})\biggl\}.
\end{equation}
In this equation, since $\tilde{X}_t$ is essentially defined as the expectation measure over $\tilde{P}$, $X_t$ can be replaced with this. The resulting expression with this expectation measure is denoted as
\begin{equation}\label{hybrid_min_x_tilde_t}
    \mathcal{J}_{\tilde{X}_t} = \sum_{j=0}^{N_j}\biggl\{\int_{t_j^+}^{t_{j + 1}^-}\frac{1}{2 \epsilon }\lvert \lvert u\rvert \rvert^2dt - \mathcal{L}^j(t, \tilde{X}_t, u) + KL(\pi^j_{0} \parallel \nu^j_{0})\biggl\}.
\end{equation}
Finally, this makes the similarities between smoothing and optimal control evident, by comparing Equation~\eqref{hybrid_oc} and Equation~\eqref{hybrid_min_x_tilde_t}
\begin{equation}\label{hybrid_oc_filter_sim}
    \min\limits_{u}\Esp \sum_{j=0}^{N_j}\biggl\{\int_{t_j^+}^{t_{j + 1}^-} g(t, X_t) dt + \Psi_T(X_t^{j})\biggl\} + \Esp \sum_{j=0}^{N_j}\biggl\{\int_{t_j^+}^{t_{j + 1}^-}\frac{1}{2 \epsilon}\lvert \lvert u_t\rvert \rvert^2 \biggl\} \sim \min_{u, \pi_0} \Esp\biggl\{\mathcal{J}_{\tilde{X}_t}\biggl\}.
\end{equation}
Hence, it can be seen that if written in terms of $g(t, x)dt$ and $\Psi_T(x)$, this yields a similar result to Equation~\eqref{smooth_oc_filtering_sub}:
\begin{equation}\label{smoothing_oc_sub}
    \begin{aligned}
        &g(t, x)dt = \frac{1}{2 \sigma_B^2}\lvert \lvert H^j(t, x)\rvert \rvert^2dt + \frac{1}{\sigma_B^2}Y_t^jdH^j(t, x), \\
        &\Psi_T(x) = -\frac{1}{\sigma_B^2}Y_T^jH^j(T, x).
    \end{aligned}
\end{equation}
Re substituting to write into a more concise form as a function of this running and terminal cost, and by expressing the sum of the running and control cost as was done in Equation~\eqref{control_running_cost_simp}, yields
\begin{equation}\label{smoothing_oc}
    \mathcal{J}_{\tilde{X}_{ot}} = \sum_{j=0}^{N_j}\biggl\{\int_{t_j^+}^{t_{j + 1}^-} l^j(t, X_t, u_t) \\
    + \Psi_T(X_t) + KL(\pi^j_{0} \parallel \nu^j_{0})\biggl\}.
\end{equation}
The only major difference between a finite-horizon stochastic hybrid optimal control problem in Equation~\eqref{hybrid_oc}, and Equation~\eqref{smoothing_oc} is the extra term $KL(\pi_{j, 0} \parallel \nu_{j, 0})$.
Now in the case that a mode mismatch exists between $Y_t^j$ and $X_t$, whereas one of the following conditions is met
\begin{equation}
    \begin{aligned}
        &Y_t \in I_{j}, X_t \in I_{j + 1} \\
        &X_t \in I_{j}, Y_t \in I_{j + 1}
    \end{aligned}.
\end{equation}
This can be problematic, as several of the terms between $Y_t$ and $X_t$ depend on them being in the same mode. This was also discussed in \cite{Rijnen2015} and \cite{Rijnen2017}, where it affected the feedback control law performance, similar to here.
Reference extensions are used to handle this case, as was done in \cite{yu2024path}, by dividing up the trajectory into two discretizations, each of which keep continuity:
$[t_j^-,..., t_{N_F}]$ and $[t_1...t_{N_b}, t_{j + 1}^+]$, for the pre- and post- contact states. To compute the $Y_j dH_t$ terms, we use for the previous mode
\begin{equation}\label{ref_ext_1}
    x_{j_{N_f}} \triangleq x(t_j^-) \cup \{Y_i^j\}_{i=1}^{N_f}, \forall i = 1...N_f,
\end{equation}
and likewise for the subsequent mode,
\begin{equation}\label{ref_ext_2}
    x_{{j + 1}_{N_b}} \triangleq x(t_{j + 1}^+) \cup \{Y_i^{j + 1}\}_{i=1}^{N_b}, \forall i = 1...N_b.
\end{equation} is used.
This way, continuity is maintained within that interval of transition, and thus avoids the comparison of a transitioned mode with that of an untransitioned mode. They are implemented at the end of the time iteration in Algorithm \ref{particle_update}.
\subsection{Hybrid Path Integral Smoothing and Filtering}
As was similar to \cite{ZHANG2023110894}, Equation~\eqref{smoothing_oc} is different compared to Equation~\eqref{hybrid_oc} from only the KL divergence term. In \cite{yu2024path}, the authors found that the optimal controller at time t, for Equation~\eqref{hybrid_oc} could be expressed as (while ignoring the noise intensity terms, as those don't exist in our model)
\begin{equation}\label{hybrid_oc_sol}
    u_t^* = \frac{\Esp_{\Psp^0}\left[\Delta W_t \exp\left(\mathcal{L}_H(t)\right)\right]}{\Delta t \times \Esp_{\Psp^0}\left[\exp \left(\mathcal{L}_H(t)\right)\right]},
\end{equation}
where $\mathcal{L}_H(t)$ is defined as the piecewise state cost:
\begin{equation}\label{L_H_meaning}
    \mathcal{L}_H(t) \triangleq \mathcal{I}_L\left[t, t^-_{j_m - 1}\right] + \sum_{j = j_m}^{N_j}\mathcal{I}_L\left[t_j^+, t_{j + 1}^-\right],
\end{equation}
and $j_m$ is the minimum $j$ that $t_j^+ \geq t$ and $\mathcal{I}_L$ is defined as
\begin{equation}\label{I_L_meaning}
    \mathcal{I}_L[a, b] \triangleq \int_a^b g(t, X_t)dt + \Psi_T(X_T^{N_j}).
\end{equation}
It can be observed that this optimal controller is independent of the distribution $\pi_{j, 0}$. Plugging Eq. [\ref{hybrid_oc_sol}] into [\ref{smoothing_oc}], we arrive at the optimization over $\pi_{j, 0}$. Now, the following substitution is used to express the optimization over the initial distributions $\pi_{j, 0}$:
\begin{equation}\label{V_t_hybrid_sub}
    V_t(x) = -\log \phi(t, x),
\end{equation}
where $\phi(t, x)$ is defined as
\begin{equation}\label{phi_t_hybrid_sub}
    \phi(t, x) = \Esp_{\Psp^0}[\exp(\mathcal{L}_H(t))].
\end{equation}
By splitting up the terms and applying properties of KL divergence, this causes our Eq. to simplify down to the following optimization problem which seeks to minimize the set of distributions $\pi_0 := \{\pi_{j, 0}, \pi_{j + 1, 0},... \pi_{J, 0}\}$
\begin{equation}\label{pi_0_optimization}
    \min\limits_{\pi_0}\Esp\biggl\{\sum_{j=0}^{N_J}  -\log{\phi(0, X^j_{0})} -\log{\nu^j_{0}(X^j_{0})} + \log\pi^j_{0}(X^j_{0})\biggl\}.
\end{equation}
The controllable stochastic model can now be plugged in to run the rollout trajectories.$X_t^j$
\begin{equation}\label{rollout_stoch_hybrid}
    dX_t^j = F_j(t, X_t^j)dt + \sigma^j(t, X_t^j)(u^j(t, X_t^j)dt + \sqrt{\epsilon}dW_t^j), \quad X_{0_j}^k \sim \pi_{0_{j}}^*,
\end{equation}
into Equation~\eqref{pi_0_optimization}, completing our optimal control setup.
\subsection{Hybrid Particle Filtering}
Now, we can apply the principles of particle filtering to this hybrid smoothing problem, analogous to how \cite{ZHANG2023110894} did this with continuous-time systems. First, the general equation for the conditional distribution of the signal $X_t$, given the measurement $Y_t$ for $0 \leq t \leq T$ is
\begin{equation}\label{conditional}
    \frac{1}{K}\sum_{k = 1}^K\delta_{X_t^k}.
\end{equation}
Now, we can use Girsanov's theorem, and our standard KL divergence formula, which was also applied to continuous-time systems
\begin{equation}\label{KL-classical}
    \frac{d\mathcal{Q}^Y}{d\tilde{\mathcal{P}}} = \frac{d\mathcal{Q}^Y}{d\mathcal{P}}\frac{d\mathcal{P}}{d\tilde{P}}.
\end{equation}
By Girsanov theorem, and from Equation~\eqref{ks_posterior_hybrid}, we can come to that at a given mode, that Equation~\eqref{KL-classical} is proportional to
\begin{equation}
    \frac{d\mathcal{Q}^Y}{d{\mathcal{P}}} \frac{d\mathcal{P}}{d\tilde{\mathcal{P}}} \propto \frac{d\nu_0}{d\pi_0}\exp{[-S_u(0, T)]},
\end{equation}
where $S_u$ is the mode-specific integral defined as
\begin{equation}\label{Su_mode_spec}
    \begin{aligned}
        &S_u(t, s) = \int_t^s \frac{1}{2 \epsilon} \lvert \lvert u_{\tau} \rvert  \rvert^2 d\tau - \mathcal{L}^j(\tau, X_\tau)
          + \frac{1}{\sqrt{\epsilon}}u_{\tau}'dW_{\tau} + \frac{1}{\sigma_B^2}Y_t(t, X_t).
    \end{aligned}
\end{equation}
We now define the weights, at a particular mode as
\begin{equation}\label{weight_mode}
    w^k_j = \frac{d\nu_0}{d\pi_0}(X_0^k)\exp[-S_u^k(t_j^+, t_{j + 1}^-)].
\end{equation}
In the ``Algorithm and Implementation'' section, we discuss how we update the weights at the start of each new mode $\mathcal{I}_j$. 
From here, you can approximate $d\mathcal{Q}^Y$. An empirical distribution is an estimate of the cumulative distribution function that generated the points in a sample. We can now combine the ideas used by the empirical distribution formula given by Equation~\eqref{KL-classical}, and use it to get the expression
\begin{equation}\label{dQY_summation}
    d\mathcal{Q}^Y \approx \sum_{k = 1}^K \hat{w}^k_j\delta_{x_{j}^k}.
\end{equation}
Here, $\hat{w}^k$ is a normalized weight, a key factor to ensure that the state estimation stays effective, s.t.
\begin{equation*}\label{normalized_distribution_weights}
    \hat{w}^k_j = \frac{w^k_j}{\sum_{k=1}^K w^k_j}.
\end{equation*}
One can view the above expression as a weighted average that aims to approximate the posterior. Therefore, the posterior distribution for $X_t^j$ for $t$ s.t. $t_j^+ \leq t \leq t_{j + 1}^-$ can be approximated by
\begin{equation}\label{posterior_approx}
    \sum_{k = 1}^K \hat{w}^k_{t_j}\delta_{X_{t_j}^k}.
\end{equation}
Consider a sliding window $[t_j^+ - H, t]$ of size $H > 0$. We can take Equation~\eqref{smoothing_oc}, and modify it with this consideration, resulting in
\begin{equation}\label{smoothing_oc_window}
    \min\limits_{u, \pi_0}\Esp \sum_{j=0}^{N_j}\biggl\{\int_{t_j^+ - H}^{t} l^j(t, X_t, u_t) + \Psi_T(X_t) + KL(\pi^j_{0} \parallel \nu^j_{0})\biggl\},
\end{equation}
The use of this interval, and where it fits into the algorithm process is shown in the loop iteration of Algorithm \ref{HPIPF}. On the other hand, the place where this setup is solved with iLQR is shown at the beginning of the per-particle iteration in Algorithm \ref{HPIPF_step}.
This sliding-window approach is used in order to reduce the computational burden of resampling the whole trajectory from the very beginning. By doing so, we avoid stacking up recursive costs over time, and keep the runtime of the algorithm consistent over each iteration. While this may initially cut performance, we discuss later how with a proper size window, they can be mitigated, as was shown in \cite{ZHANG2023110894}. 
Similarly to what was previously done in \cite{ZHANG2023110894}, we can represent these distributions as collections of weighted particles, and using the sliding mode characteristic, we arrive at the following expression for the prior distribution,
\begin{equation}
    \nu_{t - H} \approx \sum_{k = 1}^K\tilde{w}_{t - H}^k \delta_{x_{t - H}^k}.
\end{equation}
As was shown in \cite{ZHANG2023110894}, we compute a suboptimal control policy $u$ for the closed-loop dynamics. From there, the approximation of the posterior is computed  through Equation~\eqref{dQY_summation}. In other words, for prior and posterior weights, as in \cite{ZHANG2023110894}, we use the related following two equations (which also accounts for mode differences in a trajectory), respectively
\begin{equation*}
    \tilde{w}^k_{t_j^+ + H}\propto\exp\left[-S_u^k(t_j^+, t_{j + 1}^-)\right],
\end{equation*}
\begin{equation}\label{weight_update_Su}
    \hat{w}^k \propto \exp\left[-S_u^k( t_j^+, t - H)\right]\exp\left[-S_u^k(t - H, t_{j + 1}^-)\right] = \exp[-S_u^k(t_j^+, t_{j + 1}^-)].
\end{equation}
The processes to compute these weights are done at the end of Algorithm \ref{particle_update}.
\section{Algorithm and Implementation}
The reset map describing the mapping between pre-transition and post-transition modes $X_t^+ = R(t, X_t^-)$ is generally instant and nonlinear. Therefore, we use the Saltation Matrix, a first-order approximation for the varying effect of the transition along a nominal point \cite{kong2023saltation, Filippov1988, Munoz2019}, allowing for the preservation of continuity and differentiability. Given two modes $j$ and $k$, the Saltation Matrix is defined as
\begin{equation}
    \Xi_{jk} \triangleq \partial_xR_{jk} + \frac{F_k - \partial_x R_{jk} \cdot F_j - \partial_tR_{jk}\partial_x g_{jk}}{\partial_t g_{jk} + \partial_x g_{jk} \cdot F_j}.
\end{equation}
With it, the dynamics of the perturbed state $\delta X_t$ at hybrid transitions can be approximated as
\begin{equation}
    \delta X_t^+ \approx \Xi_{jk}\delta X_t^-.
\end{equation}
Notably, it has been proven to perform better than differentiating the reset map due to how it handles time differences upon hitting the guard condition \cite{kong2023saltation}.
This section provides details about how the PIPF algorithm was extended to hybrid systems and how the algorithm evaluates transition events. We assume constant step size discretization for this as well. 

For our optimal control approach, we use a modified form of the Salted iLQR algorithm found in \cite{kong2021} to compute our feedback gains and resulting particle trajectories. This approach employs the Saltation Matrix to handle mode transitions in the backwards pass of the iLQR algorithm, allowing for smooth differentiation when computing the feedback gains if the system crosses modes. The main differences between our algorithm and the Salted iLQR algorithm lie in the computation of the running and terminal cost derivatives, as our problem is slightly different.

We also assume stochasticity in the modes, namely since the ground truth mode at a given point in time is not known, each particle may not be in the same mode as the other. For estimation purposes under those conditions, we employ a form of ``voting logic,'' where we compare which mode has a greater overall particle weight. Once the mode with the greater weight is identified, the particles within that mode have their weights recomputed (so that they still sum to one), and we compute the state estimate using those particles. This logic is carried out after Algorithm \ref{HPIPF_step} is done and trajectories are obtained, as shown in the last line of the iteration in Algorithm \ref{HPIPF}.

The overall structure of the SPIPF algorithm is illustrated in Algorithm \ref{HPIPF}. Algorithm \ref{HPIPF} specifically shows how the sliding window is computed (after which Algorithm \ref{HPIPF_step} is ran), and the final state and mode estimates are computed. We show the subroutine for an incoming measurement (one step of SPIPF step) over a time window in Algorithm \ref{HPIPF_step}. It shows how we compute the feedback and feedforward gains for a reference trajectory, when we run Algorithm \ref{particle_update} using those gains, and how we get the values that we then feed back into Algorithm \ref{HPIPF}. It also uses normalization, which is carried out when needed so that all weights sum to 1 (which isn't guaranteed by our earlier weight computations).
Lastly, we show an individual particle's movement is propagated from a set of computed feedback and feedforward gains in Algorithm \ref{particle_update}. This algorithm shows how the reference extensions are used if there is a mode mismatch (the mode of the reference and the rolled-out trajectory do not align). Next, it shows how if a guard condition gets hit, the particle changes its current mode. Finally, it discusses how the new set of prior and filtered weights are computed, after which the trajectories, modes, and weights are sent to Algorithm \ref{HPIPF_step}.

\begin{algorithm}[H]
\caption{Salted Path Integral Particle Filtering (SPIPF)}
\begin{algorithmic}[0]
    \State \textbf{Input:} Total Number of Steps, Sliding Window and Sampling Frequency: $(L,H, dt)$ 
    \State Sample Paths from Prior Distribution \( \nu_0 \): $\{{\tilde{x}}_p^k\}$
    \State Weight of Prior Samples, Uniform: $\{\tilde{w}_p^k\}$
    \State Modes in Prior Distribution: $\{\tilde{d}_p^k\}$ 
    \State \textbf{Output:} State Trajectory Estimate, Mode Estimate ($\bar{x}, \bar{d}$)
    \For{$j = 1, \dots, L$}
        \State $j$-th time interval $\gets [\max(0, t_j - H), t_j]$
        \State $\{\hat{x}_j\}, \{\hat{w}_j\}, \{\hat{d}_j\}, \{\tilde{x}_{p, j}\}, \{\tilde{w}_{p, j}\}, \{\tilde{d}_{p, j}\}$ $\gets$ Algorithm 2 for $j$-th time interval
        \State $\bar{x}, \bar{d}$ $\gets$ Weighted averaging and voting logic
    \EndFor
\end{algorithmic}
\label{HPIPF}
\end{algorithm}

\begin{algorithm}[H]
\caption{One Step of SPIPF}
\begin{algorithmic}[0]
\State \textbf{Input:} Cost: $\mathcal{C} = \{g, \Psi\}$
    \State Number of Particles, Sliding Window Size, Resampling Threshold: $\Phi = \{K, H, \gamma_{thres}\}$
    \State Prior Samples, Weights, Modes:  $\mathcal{P}_{sample} = \{\tilde{x}_{p, j}, \tilde{w}_{p, j}, \tilde{d}_{p, j}\}$

\State \textbf{Output:} Filtering (Posterior) Results: $\{{\hat{x}}_j, \hat{w}_j, \hat{d}_j\}$
\State Updated Sampling (Prior) Results: $\{\tilde{x}_{p, j}, \tilde{w}_{p, j}, \tilde{d}_{p, j}\}$
\State \textit{\# Inputs for Computing Gains}: $\mathcal{X,M}$
\State $k_{feedforward}, K_{feedback}$ $\gets$ Solve Hybrid iLQR \eqref{smoothing_oc_window}

\For{$k \gets 1, \dots, K$}
    \State $\{\hat{x}^k\}, \{\hat{w}^k\}, \{\hat{d}^k\}, \{\tilde{x}_p^k\}, \{\tilde{w}_p^k\}, \{\tilde{d}_p^k\}$ $\gets$ Algorithm 3
    \State Normalize sample and filtering weights $(\tilde{w}_j, \hat{w}_j)$
\EndFor
\State \textit{\# Check resampling:}
\State Effective ratio: $\gamma = \frac{1}{K \sum_{k=1}^K (\hat{w}^k)^2}$
\If{$\gamma < \gamma_{\text{thres}}$}
    \State $\{\tilde{x}_p^k\} \sim \text{multinomial}\big(\{\tilde{x}_{t_j}^k\}, \{\tilde{w}^k \exp(-S_u^k(t_i, t_j))\}\big)$
    \State $\{\tilde{w}_p^k\} \propto \exp\big(S_u^k(t_i, t_j)\big)$
    \State $\{\tilde{d}_p^k\} \gets$ Mode $\tilde{d}^k$ at $\tilde{x}^k_p$
\EndIf
\end{algorithmic}
\label{HPIPF_step}
\end{algorithm}

\begin{algorithm}[H]
\caption{One Step of the Particle Update in SPIPF Step}
\begin{algorithmic}[0]
\State \textbf{Input:} Particle to Sample: $k$
    \State Cost: $\mathcal{C} = \{g, \Psi\}$
    \State Number of Particles, Sliding Window Size, Resampling Threshold: $\Phi = \{K, H, \gamma_{thres}\}$
    \State Prior Samples, Weights, Modes:  $\mathcal{P}_{sample} = \{\tilde{x}_{p, j}, \tilde{w}_{p, j}, \tilde{d}_{p, j}\}$
    \State Current Feedforward, Feedback Gains:  $(k_{feedforward}, K_{feedback})$
\State \textbf{Output:} Filtering (Posterior) Results: $\{\hat{x}^k_j, \hat{w}^k_j, \hat{d}^k_j\}$
\State Updated Sampling (Prior) Results: $\{\tilde{x}^k_{p}, \tilde{w}^k_{p}, \tilde{d}^k_{p}\}$
\If{ref. mode != $d_j$}
    \State Use reference extensions to propagate or backpropagate reference dynamics \eqref{ref_ext_1}, \eqref{ref_ext_2}
\EndIf
\State Compute control inputs using computed gains, sample $k$-th trajectory $X^k_j$ initialized by $X^k_p$ with \eqref{rollout_stoch_hybrid}
\While{time evolves over $[t_i, t_j]$}
        \If{$\boldsymbol{x}_k$ crosses a guard $g_{d_k d_l}(x) = 0$}
            \State Transition to domain $d_l$ using reset map 
            \State $\boldsymbol{x}_k$ $\gets R_{d_k d_l}(\boldsymbol{x}_k)$
            \State Update domain index: $d_k \gets d_l$
        \EndIf
\EndWhile
\State Evaluate the value of $S_u^k$ over the trajectory $X^k_j$, modes $d^k_{t_j}$ \eqref{Su_mode_spec}
\State Filtering samples, modes: $X^k = X^k_{t_j}$, $d^k = d^k_{t_j}$
\State Filtering weight: $\hat{w}^k_j \, \propto \, \tilde{w}_p^k \exp\big(-S_u^k(t_i, t_j)\big)$ \eqref{weight_update_Su}
\State \emph{\# Prior information for usage in next sliding window}
\State Prior sample, mode: $\tilde{x}_p^k = X_{t_i+1}^k$, $\tilde{d}_p^k = d_{t_i+1}^k$ 
\State Prior sampling weight: $\tilde{w}_p^k \, \propto \,\tilde{w}^k_{t_i} \exp\big(-S_u^k(t_i, t_i+1)\big)$ \eqref{weight_update_Su}
\end{algorithmic}
\label{particle_update}
\end{algorithm}

\section{Numerical Examples}
In this section, two numerical examples are presented to demonstrate the working of the algorithm and compare the accuracy with respect to the exact posterior distribution. Two systems are considered for this: (1) Bouncing Ball System and (2) Spring Loaded Inverted Pendulum. Each system is set up to transition exactly once approximately midway through the time sequence. By default, $dt = 0.01$, $K = 50$, $H = 10$, $\epsilon = 0.1$, while for SLIP, the defaults are $dt = 0.001$, $K = 100$, $\epsilon = 0.01$, and we test each system across 50 trials. Where $K$ denotes the number of particles, $H$ is the sliding window, $dt$ is the time step and $\epsilon $ is the noise scaling factor. This is tested using Gaussian noise as there is a closed-form known posterior (the SKF), under the expectation that a particle filter's inherent flexibility will allow performance to translate.\\
All the systems considered have at least one hybrid event in their dynamics (of which the timing is unknown to the algorithm). The first step is to calculate the true posterior of the systems. The SKF \cite{KongKalman} was used to estimate the true posterior, and the system is simulated with Gaussian noise (under the expectation that performance can hence be extended to non-Gaussian noise, which may not have closed-form posteriors). The results from the posterior estimate using SPIPF were then compared against this and the time-averaged mean squared error (which is referred to as Mean MSE) was calculated. Experiments were conducted based on (a) Number of particles used for simulation $K$, (b) the time discretization $dt$, and (c) the sliding window size $H$ (which were found to be contributing factors in performance from \cite{ZHANG2023110894} and \cite{Chen2022}).  \\
The Mean MSE was defined as the following additive expression
\begin{equation}
    \text{Mean MSE} = \frac{1}{N} \sum_{n=1}^{N} \left( (x(t_n) - \hat{x}(t_n))^T (x(t_n) - \hat{x}(t_n)) \right),
\end{equation}
where \( N \) is the number of time steps, \( \hat{x}(t_n) \) is the state
estimate at time \( t_n \), and \( x(t_n) \) is the true state at time
\( t_n \). For each measurement noise, process noise, and time
step combination, the filter is run on randomly sampled starting conditions with randomly sampled process noise and randomly sampled measurements.

The algorithm is tested against two baselines: a Sequential Importance Resampling (SIR) Multi-Mode Particle Filter (employing the same previously mentioned voting logic for deciding particle mode), as well as a modified form of our algorithm that uses 0-control (SPIPF-0). This ablation (0-control) is presented in comparison to the hybrid pipf algorithm in the results below. MATLAB's Parallel Computing toolbox was used to carry out multiple simulations of our algorithm and baselines simultaneously on fixed sets of dynamics (which were generated prior to each run).

\subsection{Bouncing Ball System}

A 1-D bouncing ball was also simulated under elastic impact \cite{Hybridsystems}. This example can be viewed as a simplified and structurally homogeneous case of the Tensegrity robot \cite{Garanger2021SoftTensegrity} or other landing systems. States that were considered were the position and velocity. Guard sets were defined with respect to the velocity such that the domain of  \( F_1 \) is when the velocity is negative (falling down) and the domain of  \( F_2 \) is when the velocity is positive (bouncing back up). The guard conditions here are defined as

\begin{equation}
    \dot{x} > 0 \quad \text{and} \quad \dot{x} < 0, \\
    \text{where } \mathbf{x} = \begin{bmatrix} x_1 \\ x_2 \end{bmatrix} = \begin{bmatrix} x \\ \dot{x} \end{bmatrix}.
\end{equation}
A visualization of this system is shown below in Fig. \ref{bb_viz}.
\begin{figure}[H]
    \centering
    \includegraphics[width=0.50\linewidth]{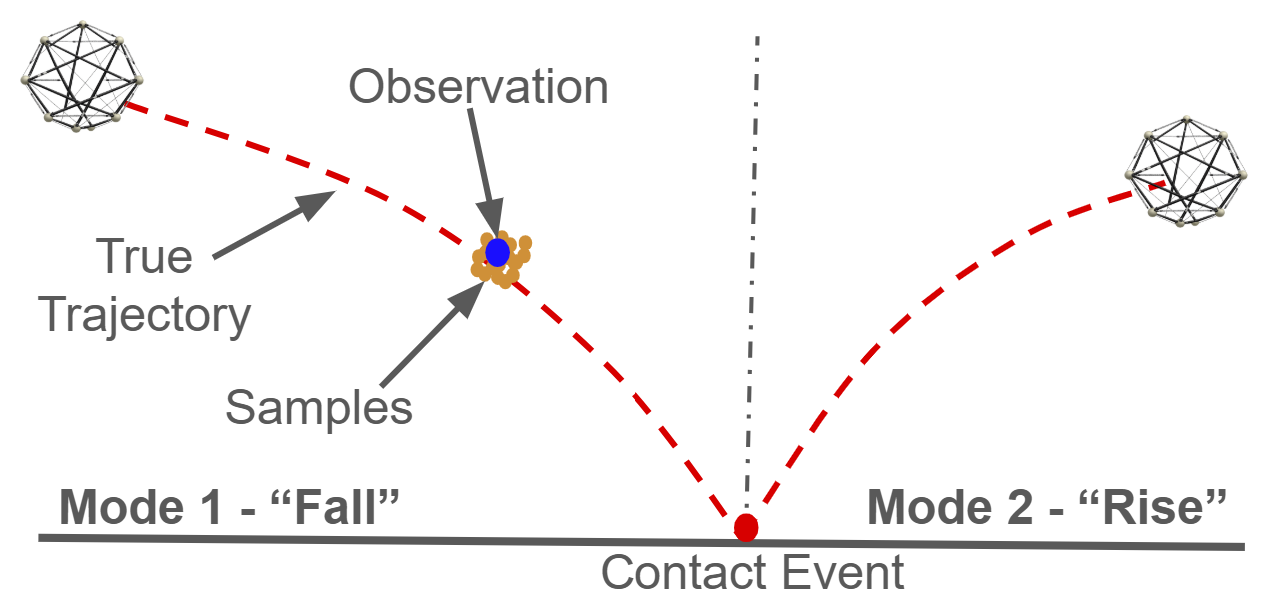}
    \caption{Bouncing Ball Visualization}
    \label{bb_viz}
\end{figure}
Now, the dynamics for each mode required a "simulated" control input in this case as this is essentially solving an iLQR problem for each time window of the PIPF algorithm. So the dynamics are given as

\begin{equation}
    F_1(x, u) = F_2(x, u) := 
    \begin{bmatrix}
        \dot{x} \\
        \frac{u - mg}{m}
    \end{bmatrix}+ \sqrt{\epsilon} 
\begin{bmatrix}
0 \\
\frac{1}{m}
\end{bmatrix}
dW_t^{(1,2)}.
\end{equation}

Hybrid mode 1 transitions to 2 when the ball hits the ground, \( g_{1,2}(x) := x \), and mode 2 transitions to 1 at \( g_{2,1}(x) := \dot{x} \). When mode 1 transitions to 2, an elastic impact is applied, with a coefficient of restitution\( -e\). The reset map from 2 to 1 is identity.

The Jacobian of the reset map and saltation matrix are
\begin{equation}
    D_x R_{1,2} =
    \begin{bmatrix}
        1 & 0 \\
        0 & -e
    \end{bmatrix}, \quad
    \Xi_{1,2} =
    \begin{bmatrix}
        1 & 0 \\
        \frac{-e (u - mg)(e + 1)}{m \dot{z}^2} & -e
    \end{bmatrix}.
\end{equation}
\subsection{ Spring Loaded Inverted Pendulum (SLIP) System}
The last system considered is the SLIP System, which is used to model locomotion of bipedal or quadrupedal robots.  This system was included and useful to analyze since it involves both nonlinear dynamics and non-identity reset maps. The system is established using point mass $m$, and leg with a spring constant $k$ and original length $r_0$. The following is the overall system's state vector $\{p_x, v_x, p_z, v_z, \theta, \dot{\theta}, r, \dot{r}\}$. Here, $(p_x, p_z)$ are the horizontal and vertical positions of the body, $(v_x, v_z)$ are the respective velocities, $\theta, \dot{\theta}$ are the angle and its respective velocity component between the body and the ground, and $r, \dot{r}$ are the leg length and the changing rate. 

This system consists of two modes, $\mathcal{I} = \{I_1, I_2\}$, where $I_1$ is the \emph{flight mode} $I_1 \triangleq \{X_t^1 \rvert p_z - r_0 \sin\theta \geq 0\}$, and $I_2$ is the \emph{stance} mode $I_2 \triangleq \{X_t^2 \rvert p_z - r_0 \sin \theta < 0\}$. Since a polar setup is used for the stance mode, there are different state spaces for each of the two systems. The system is visualised below in Fig. \ref{slip_viz}:
\begin{figure}[H]
    \centering
    \includegraphics[width=0.50\linewidth]{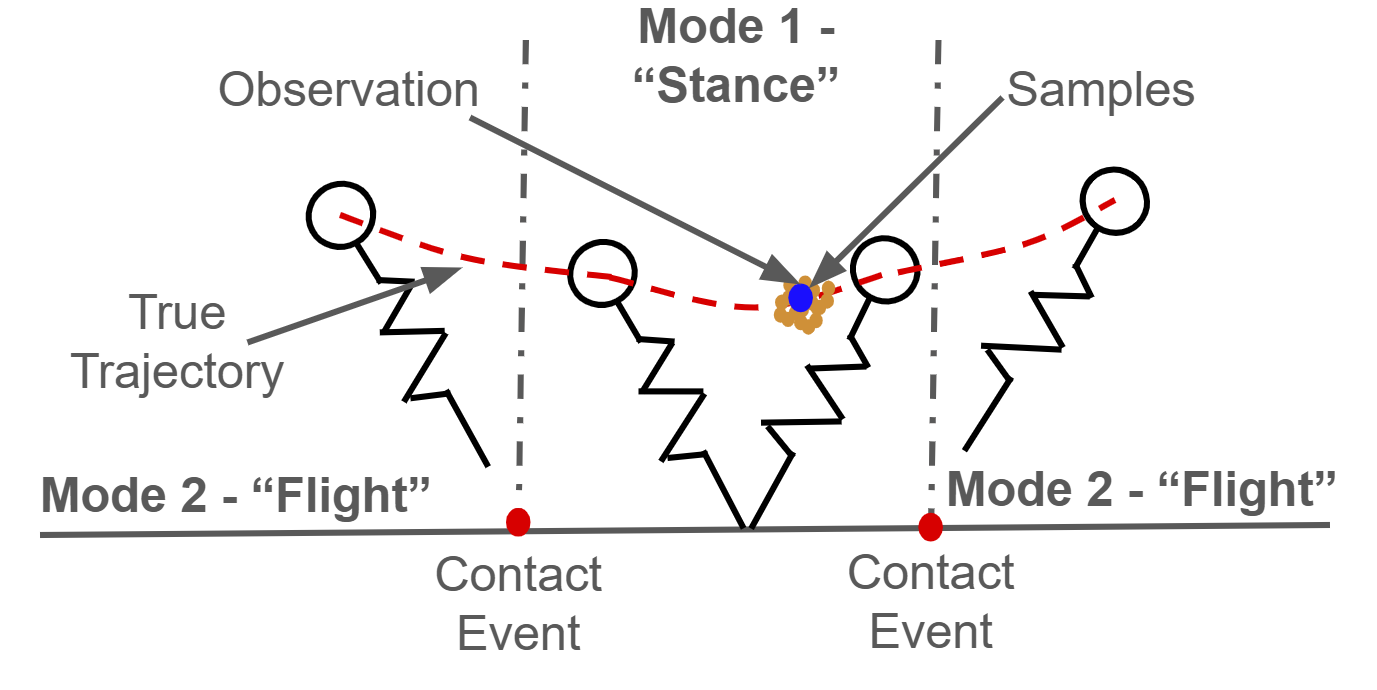}
    \caption{SLIP Visualization}
    \label{slip_viz}
\end{figure}
In the flight mode, the state space and smooth flow are defined as follows:
\begin{equation}\label{flight_mode_dyn}
    \begin{aligned}
        &X_t^1 = \begin{bmatrix}
            p_x & v_x & p_z & v_z & \theta
        \end{bmatrix}, \\
        &dX_t^1 = \left(\begin{bmatrix}
            \dot{v_x} \cr 0 \cr \dot{v}_z \cr -g \cr 0
        \end{bmatrix} + \begin{bmatrix}
            0 \cr u^1_{1, t} \cr 0 \cr u^1_{2, t} \cr u^1_{3, t}
        \end{bmatrix}\right)dt + \begin{bmatrix}
            0 & 0 & 0 \cr 1 & 0 & 0 \cr 0 & 0 & 0 \cr 0 & 1 & 0 \cr 0 & 0 & 1
        \end{bmatrix}dW_t^1,
    \end{aligned}
\end{equation}
where $g$ is the gravity, and it is assumed that the system has control of its body and leg velocities.
On the other hand, in the stance mode, the state space and smooth flow are defined as
\begin{equation}\label{stance_mode_dyn}
    \begin{aligned}
        &X_t^2 = \begin{bmatrix}
            \theta & \dot{\theta} & r & \dot{r}
        \end{bmatrix}, \\
        &dX_t^2 = \left(\begin{bmatrix}
            \dot{\theta} \cr \frac{-3 \dot{\theta} \times \dot{r} - g \times \cos \theta}{r} \cr \dot{r} \cr \frac{k(r_0 - r)}{m} - g\sin \theta + \dot{\theta}^2r
        \end{bmatrix} + \begin{bmatrix}
            0 \cr 0 \cr \frac{m}{r^2}u^2_{1, t} \cr \frac{k}{m}u^2_{2, t}
        \end{bmatrix}\right)dt + \begin{bmatrix}
            0 & 0 \cr 0 & 0 \cr \frac{m}{r^2} & 0 \cr 0 & \frac{k}{m}
        \end{bmatrix}\begin{bmatrix}
            dW^2_{1, t} \cr dW^2_{2, t}
        \end{bmatrix}.
    \end{aligned}
\end{equation}
The reset maps between the two modes is defined as
\begin{subequations}
    \begin{equation}\label{reset_2+_flight}
        X^2_{t^+} = R_{12}(X^1_{t^-}) = \begin{bmatrix}
            \theta^- \cr \frac{p_x^- v_z^- - p_z^- v_x^-}{r_0^2} \cr r_0 \cr -v_x^- \cos \theta^- + v_z^-\sin \theta^-
        \end{bmatrix},
    \end{equation}
    \begin{equation}\label{reset_1-_stance}
        X^1_{t^+} = R_{21}(X^2_{t^-}) = \begin{bmatrix}
            p^-_{x, T} + r_0 \cos \theta^- \cr 
            \dot{r}^- \cos \theta^- - r^-\dot{\theta}^- \sin  \theta^- \cr 
            r_0 \sin \theta^- \cr 
            r_0 \dot{\theta}^- \cos \theta^- + \dot{r}^- \sin \theta^- \cr \theta^-
        \end{bmatrix},
    \end{equation}
\end{subequations}
where $p_{x, T}$ is assumed as the pole x-position, and goes unchanged in the stance mode. In the monopedal hopping quadcopter developed in \cite{Bai2024AnAM}, a similar set of hybrid modes (with modified dynamics) were also used.

% \subsection{Constant Flow System with Hybrid Event}

% We first considered the constant flow system defined above. The system was simulated for different time steps for 5 seconds,  \(\Delta\) = 5, 1, 0.1, and 0.01. The results were compared with the SKF results for this system. This is a 2-D system with a hybrid event at  \( x_1 = 0 \). The saltation matrix calculation for this system is mentioned in the appendix. 
\section{Results}
\subsection{Bouncing Ball}
The bouncing ball system is fairly linear, and therefore we can get solid estimation results by using a low amount of particles. In our first set of experiments, we study the impact of changing $K$ on mode prediction, MSE, and effective sample size (ESSE).
\begin{figure}[H]
    \centering
    \includegraphics[width=0.48\linewidth]{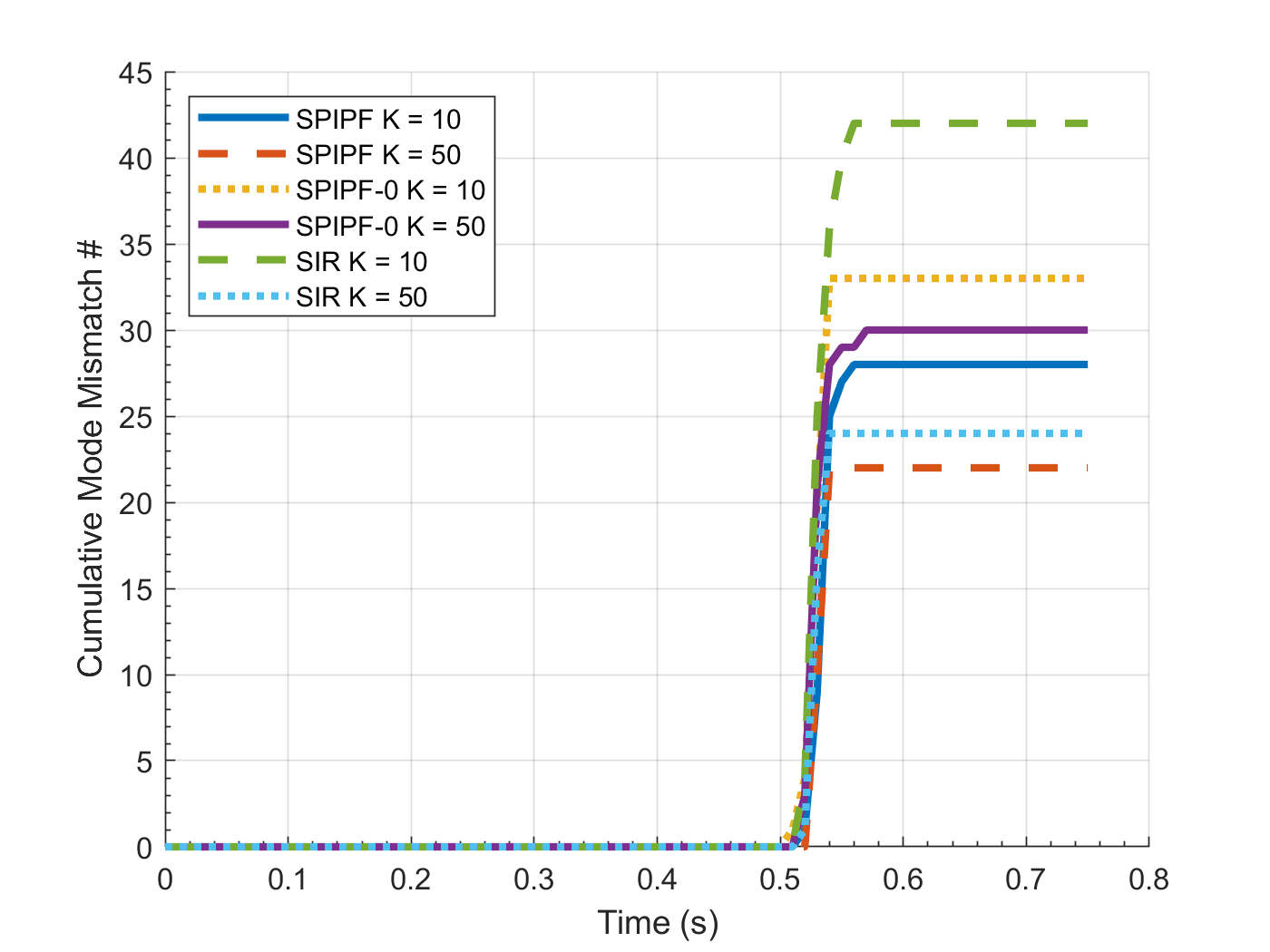}
    \centering
    \includegraphics[width=0.48\linewidth]{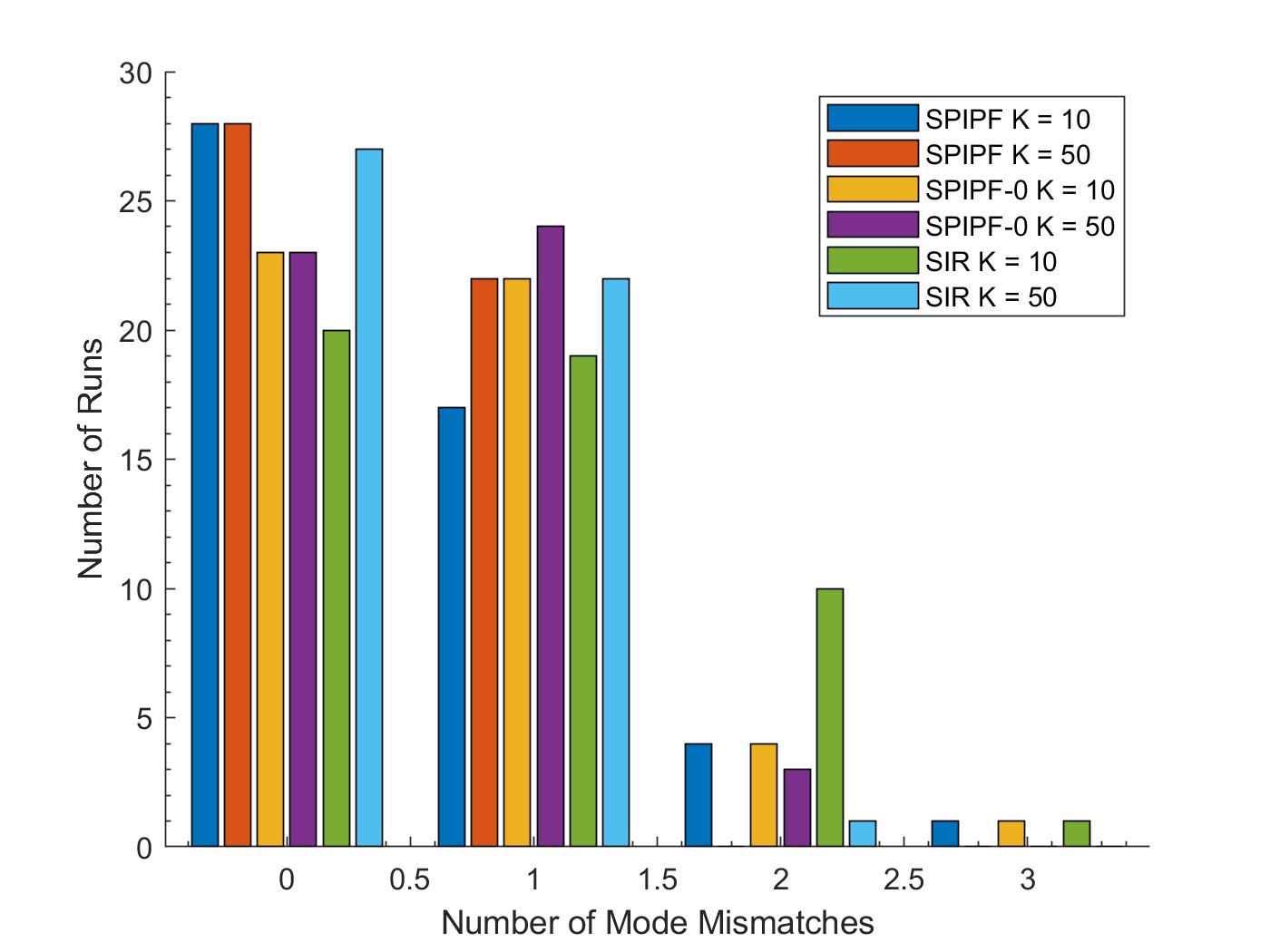}
    \centering
    \includegraphics[width=0.48\linewidth]{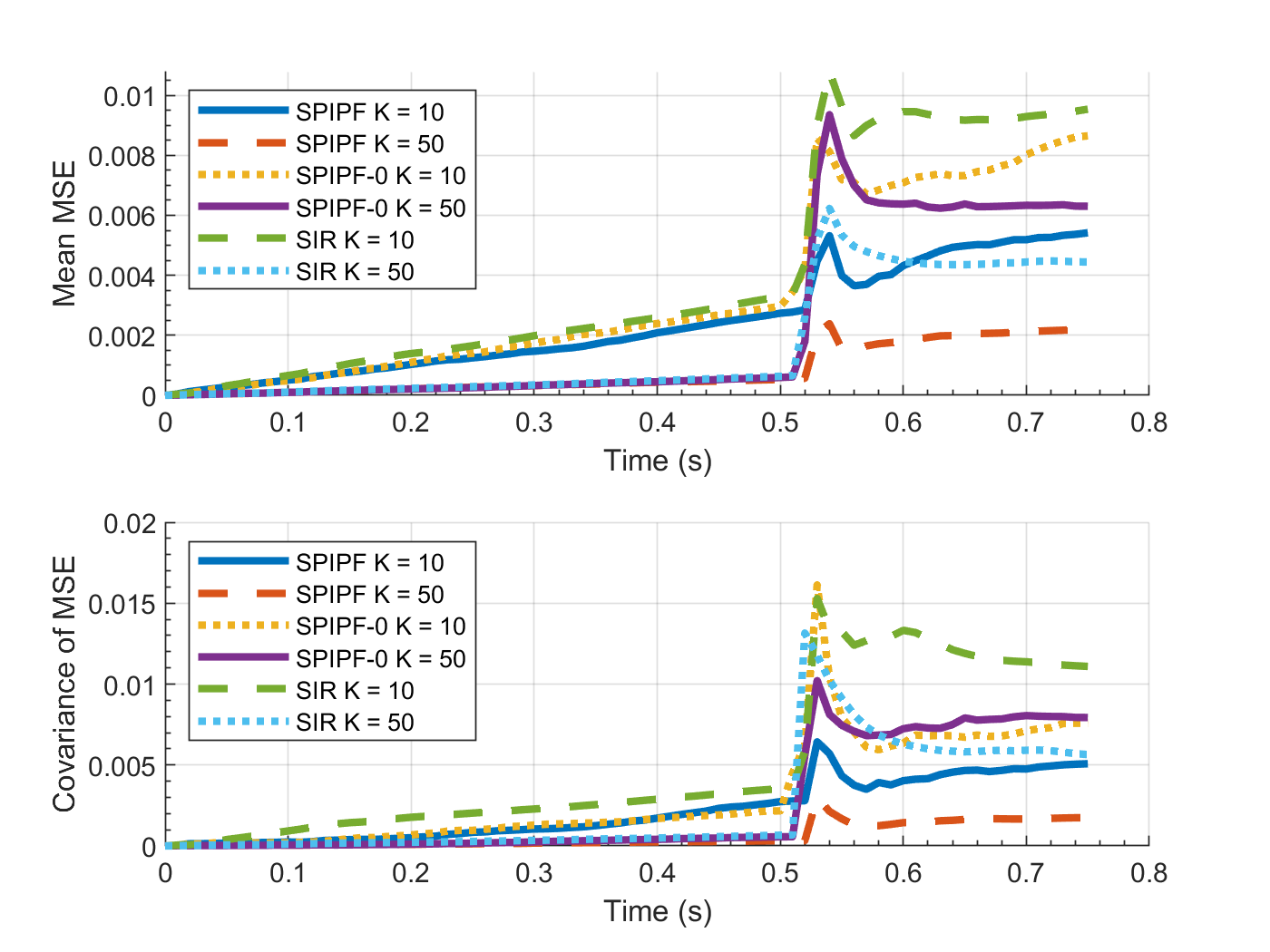}
    \centering
    \includegraphics[width=0.48\linewidth]{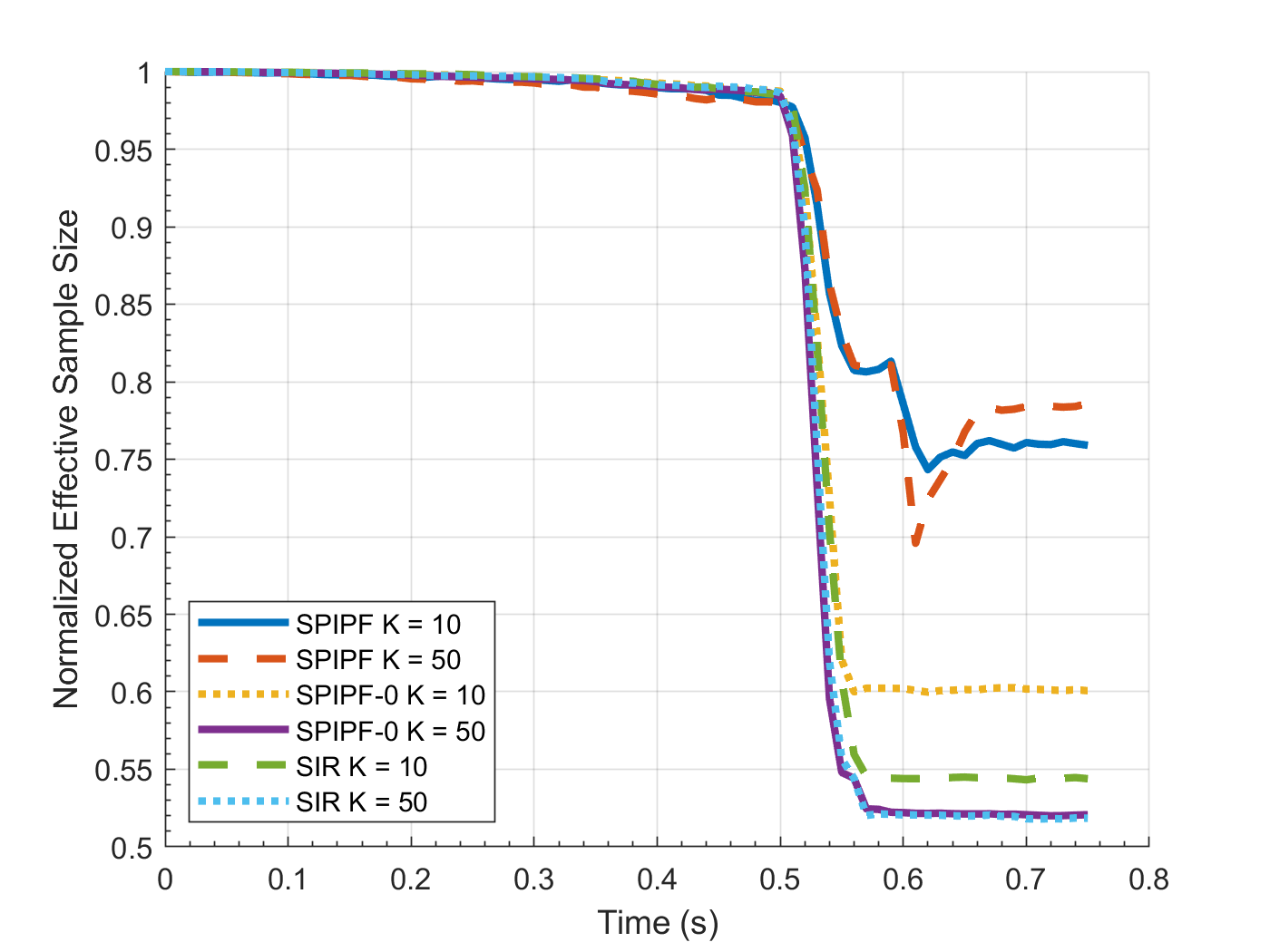}
    \caption{Results from Changing Particle Count}
    \label{bb_K_vary}
\end{figure}
As can be seen, our method (SPIPF) significantly outperforms the baselines (SPIPF-0, and SIR), for all particle counts after the transition. Notably, SPIPF with 10 particles is better than the baselines, even when they  use 50 particles. With 50 particles on SPIPF, we see higher mode prediction accuracy versus 10 particles, with an interesting tradeoff of lower ESSE. Note that the ESSE is still well-above the resampling threshold. One interesting characteristic is the dip in ESSE around 0.1 seconds after the transition. This makes sense, as per our sliding window logic, the particles at 0.1 seconds after the transition would be either initialized in mode 1 or 2, depending on the prior, which would cause a sharp deviance in particle weight at that point. On the other hand, the histogram plot involving mode mismatches is less conclusive. We believe that in the presence of stochasticity, these results are not incredibly useful to interpret, so long as the mode accuracy is within 0.2 seconds. 

Next, we vary the size of the sliding window $H$, and show the results below for 50 particles.
\begin{figure}[H]
    \centering
    \includegraphics[width=0.48\linewidth]{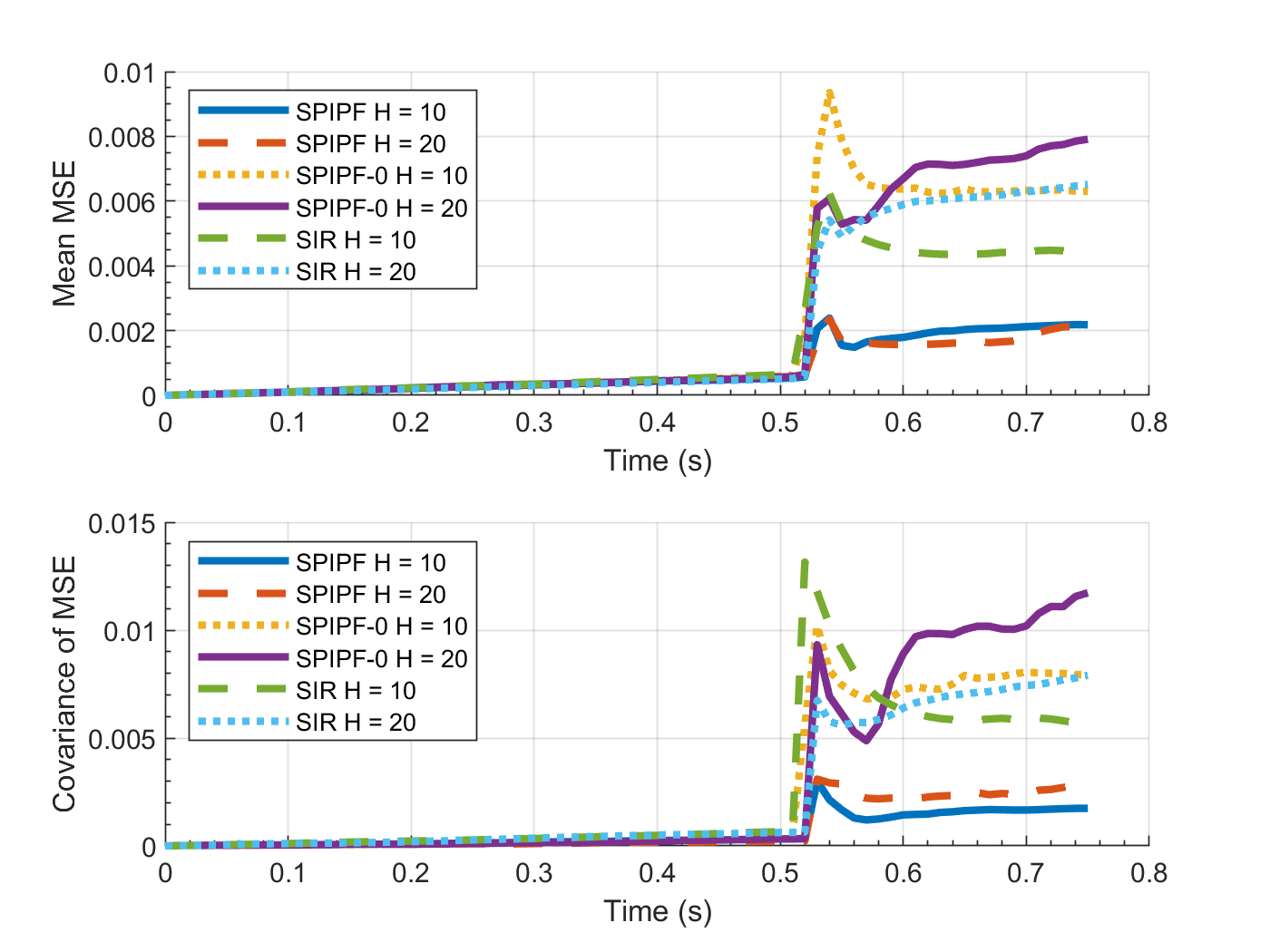}
    \centering
    \includegraphics[width=0.48\linewidth]{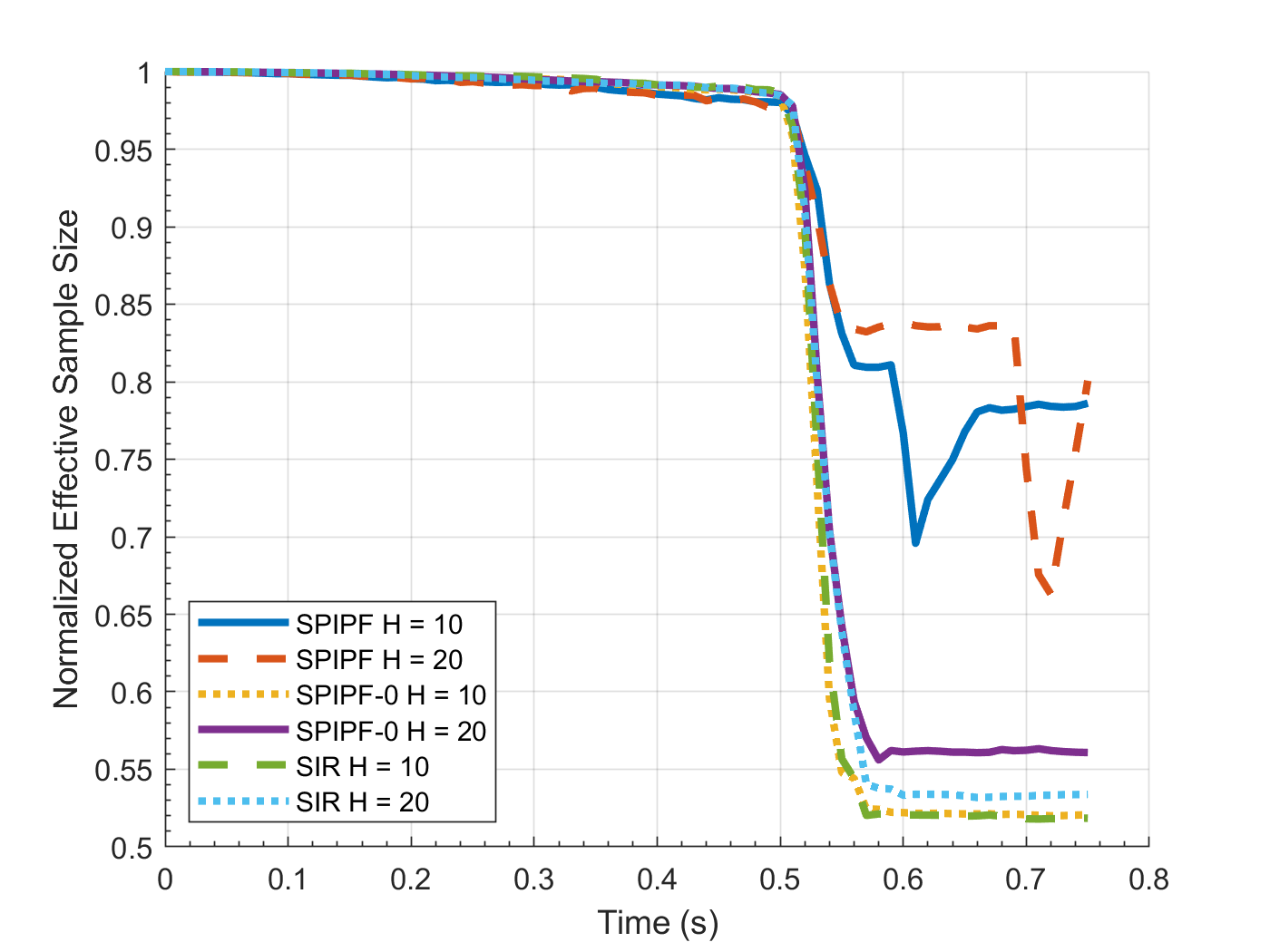}
    \caption{Results from Changing Sliding Window Size}
    \label{bb_H_vary}
\end{figure}
As expected, our model again outperforms the baseline, and with a higher sliding window size, we see more accuracy and lower covariance with respect to the MSE. This is expected as that increases the window that our algorithm looks backward. Again, we posit the same root cause for the dip in ESSE in the graph of $H = 20$, but this aligns with 0.2 seconds after transition (which again makes sense as that is 20 timesteps after the transition).

Now, we look at the results of varying the timestep $dt$.
\begin{figure}[H]
    \centering
    \includegraphics[width=0.48\linewidth]{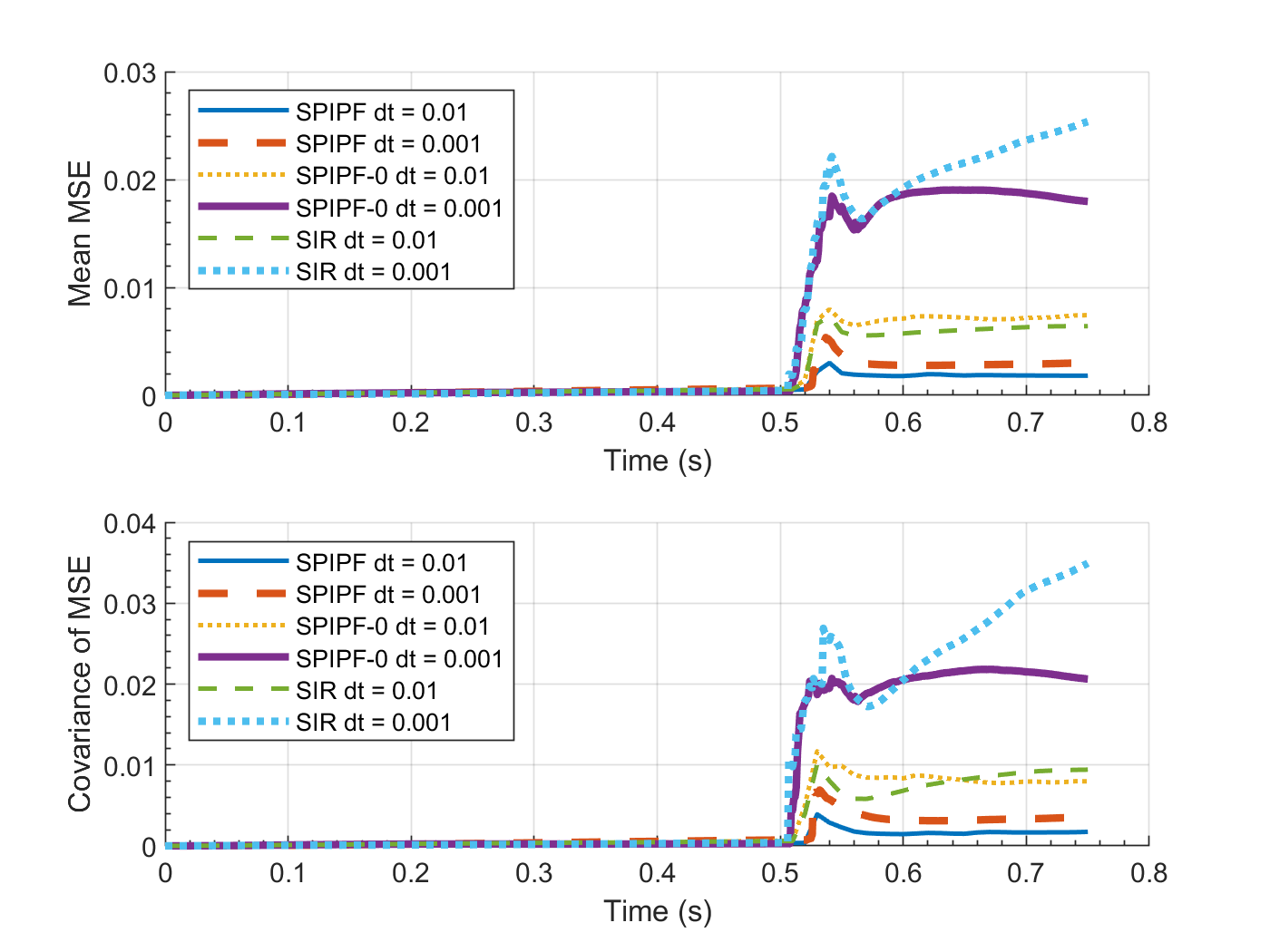}
    \centering
    \includegraphics[width=0.48\linewidth]{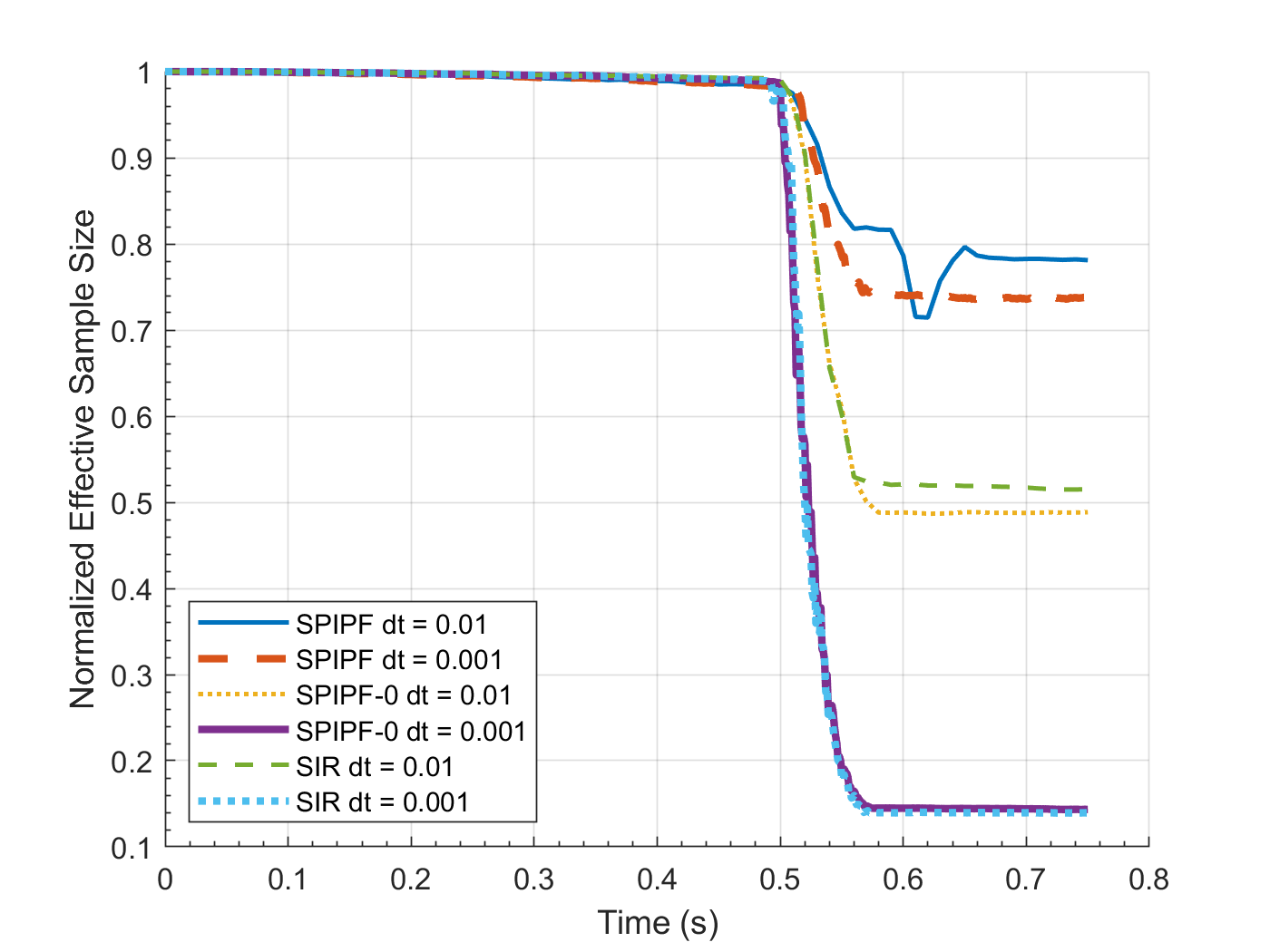}
    \caption{Results from Changing Algorithm Timestep}
    \label{bb_dt_vary}
\end{figure}
Here, with a lower algorithm timestep, we actually see SPIPF (top left graph, orange line) substantively outperforms the baselines when it comes to mode prediction. At the same time, we see the dip in ESSE we previously saw from being near a sample window almost disappearing completely, which makes sense as the smaller timestep should reduce those perturbations. At the same time, there are somewhat minimal differences in Mean MSE and Covariance of MSE, as expected due to the linearity and simplicity of our system.
\subsection{SLIP System}
Numerical simulations were carried out using MATLAB for the aforementioned systems with at least one hybrid event. 
As stated earlier, we present results for $\epsilon = 0.01, dt = 0.001$. Since we showed the relationship between algorithm timestep in the results for bouncing ball, we solely stick to the results associated with particle count and sliding window size in this analysis. We also do not allow for resampling in our method here either.
\begin{figure}[H]
    \centering
    \includegraphics[width=0.48\linewidth]{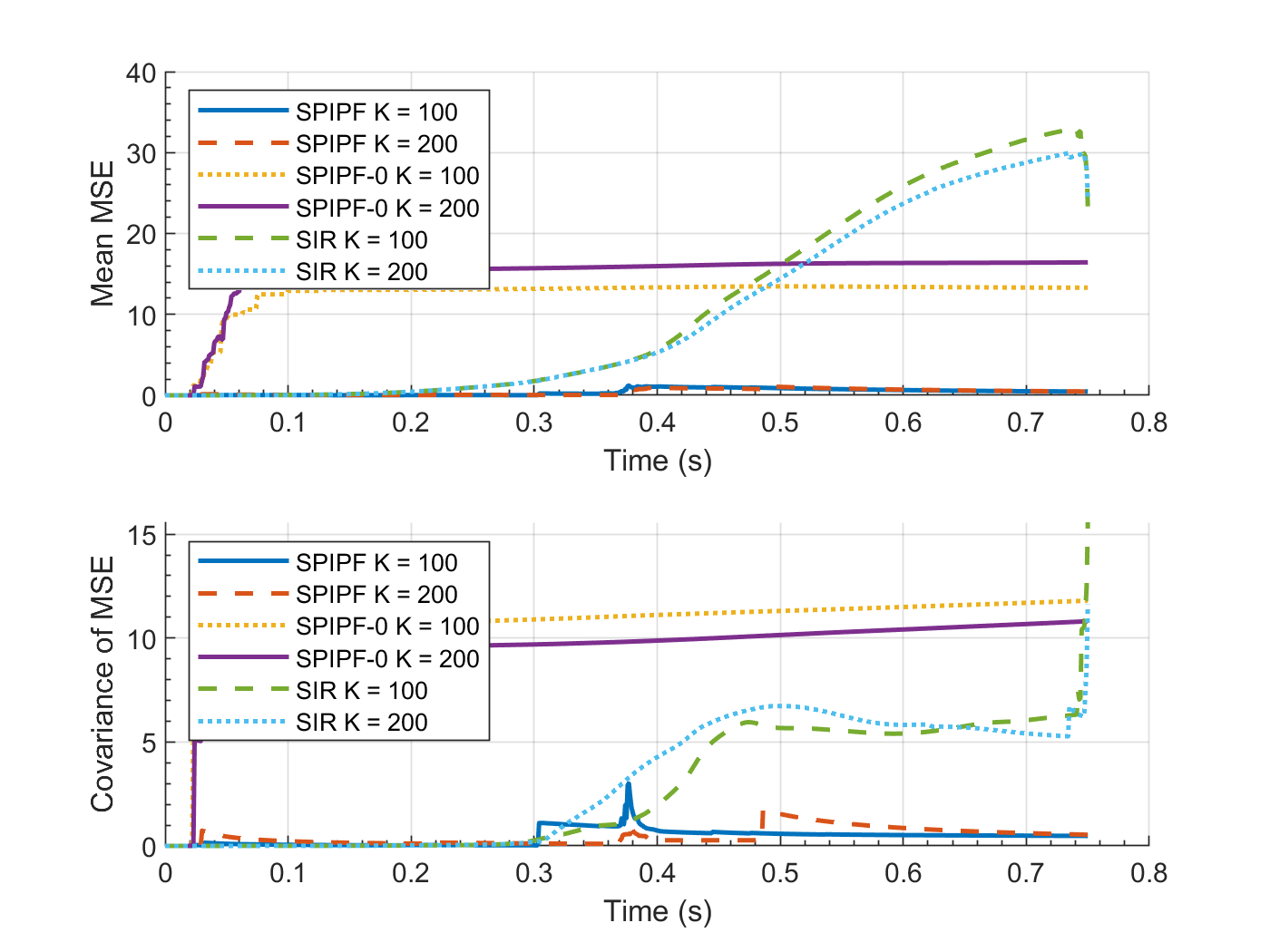}
    \centering
    \includegraphics[width=0.48\linewidth]{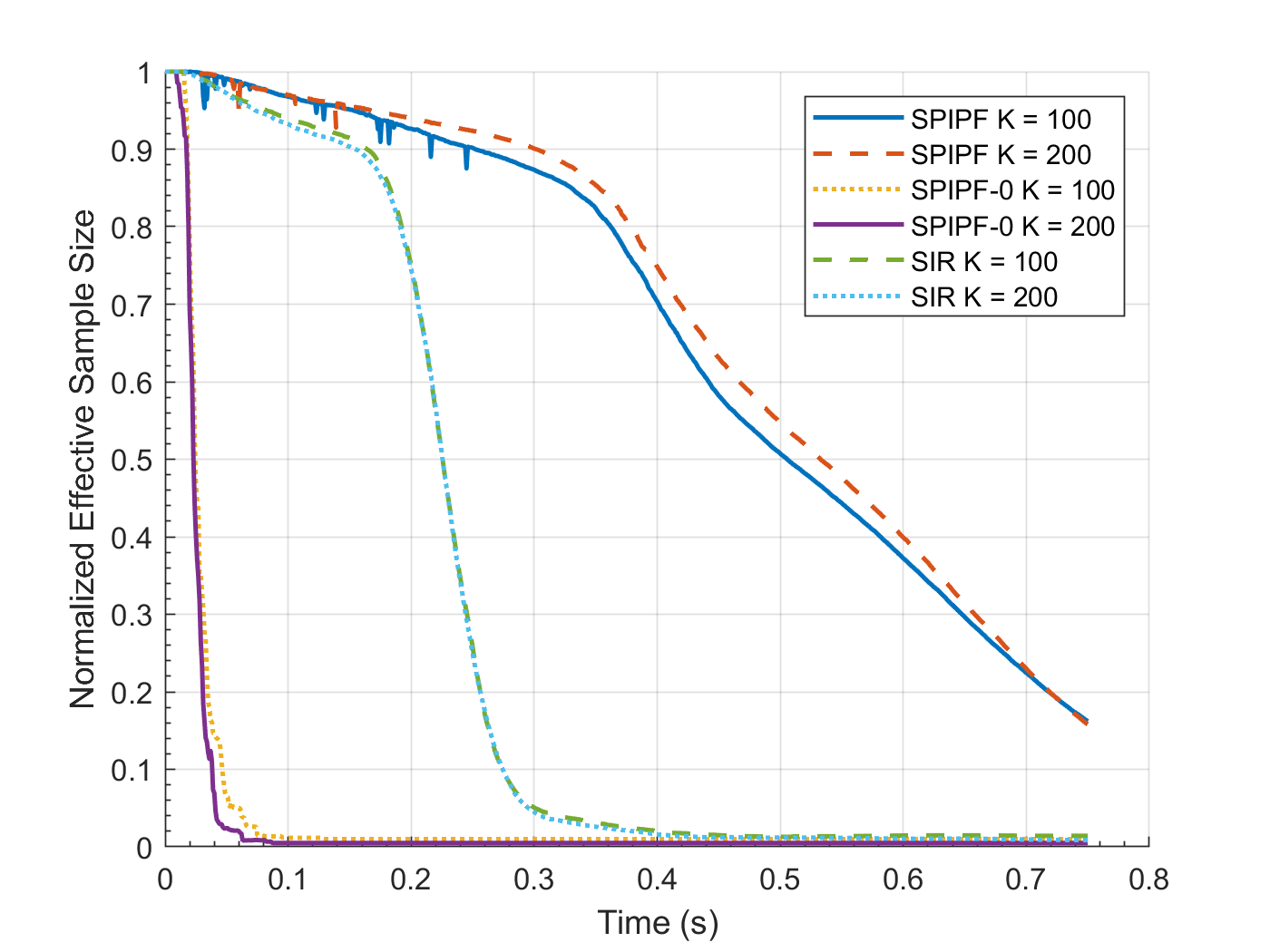}
    \caption{Results from Changing Particle Size with SLIP System}
    \label{slip_K}
\end{figure}
From the results shown in Fig. \ref{slip_K}, we can see that our algorithm outperforms the baseline methods with respect to ESSE throughout, both before and after the transition (which happens in the region of 0.35 seconds). Furthermore, we see substantively lower MSE values across both mean and covariance. As the system was highly unstable, we thresholded our results to only include the simulations that yielded below a certain MSE (which was set to 50), providing a similar number of trials per algorithm. Lastly, SPIPF also reports substantively lower MSE covariance than the baselines. 
Next, we vary the sliding window size across the trials, and obtain further results, as shown below in Fig. \ref{slip_H}.
\begin{figure}[H]
    \centering
    \includegraphics[width=0.48\linewidth]{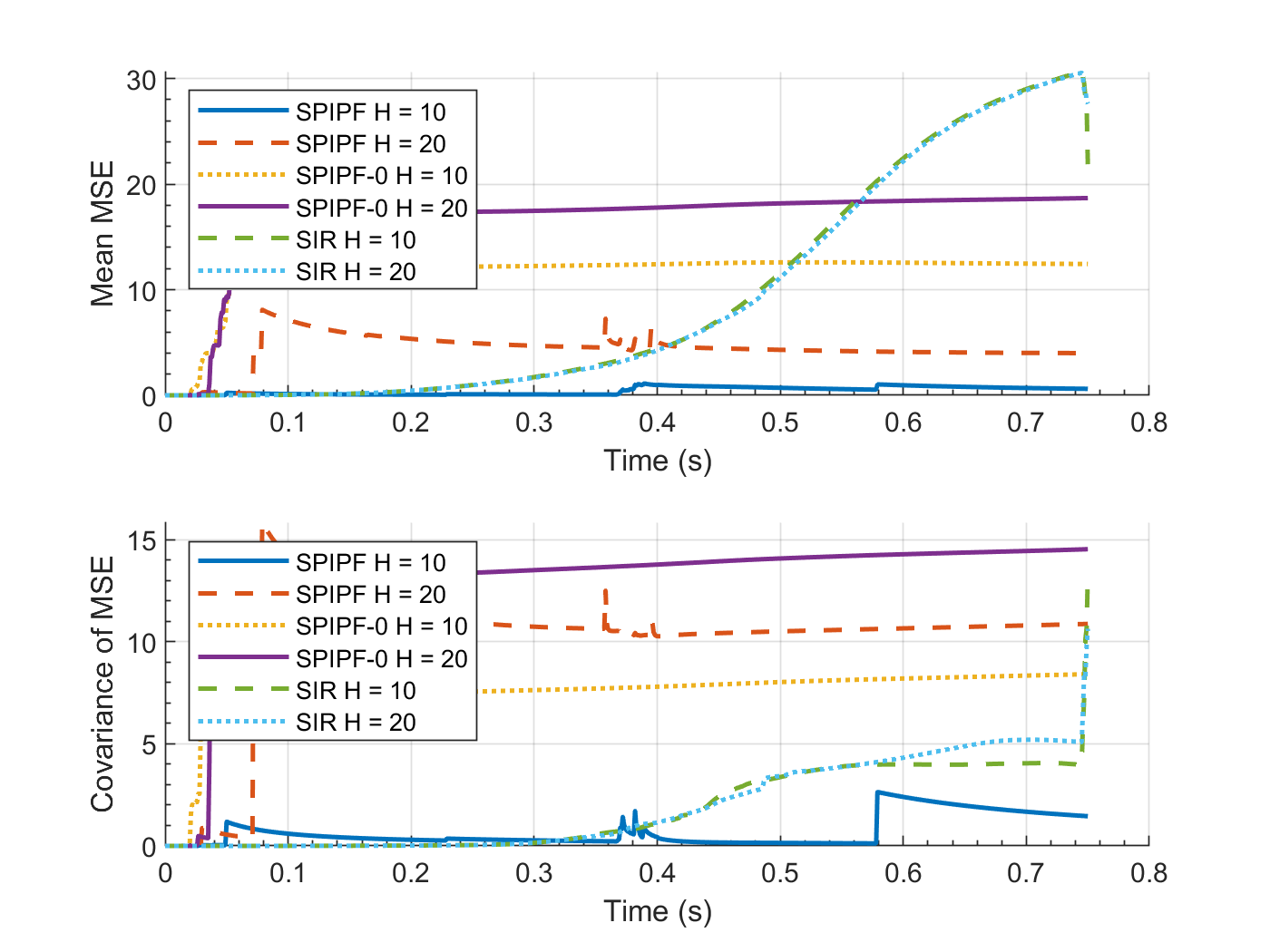}
    \centering
    \includegraphics[width=0.48\linewidth]{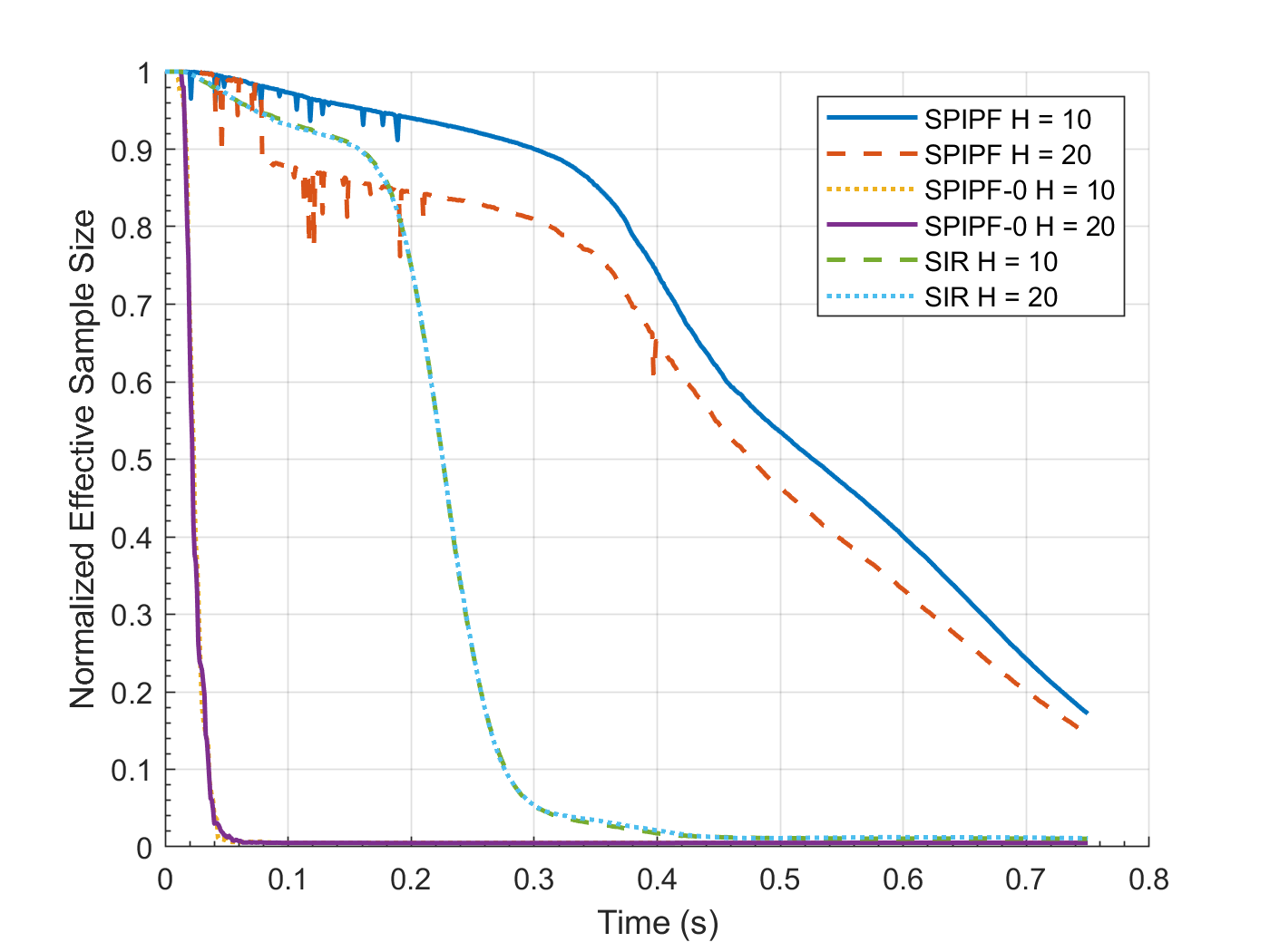}
    \caption{Results from Changing Sliding Window Size with SLIP System}
    \label{slip_H}
\end{figure}
Notably, we find it interesting that increasing the sliding window size does not necessarily yield improved performance. We believe that this has to do with the overall instability of the system, and in analysis of specific simulations, we found that SPIPF and the baselines, could predict a variety of transition times, around $\pm 0.1$ seconds before or after the actual transition occurred (which was set to be around 0.35 seconds). We believe this has to do with the instability of the system with respect to transitions, and during a longer sliding window, the system ends up triggering transitions earlier (which can be seen in the increased number of downward spikes in the effective sample size plot). This in turn suggests a high sensitivity to guard detection in the analysis of this system.
\section{Conclusion}
This paper applies path integral particle filtering to hybrid systems defined by contact using Saltation Matrices to map out uncertainty propagation and reference extensions to handle mode mismatches. Our work allows for model-agnostic non-Gaussian state estimation for hybrid systems with an optimal control approach, enabling for reduction of weight degeneracy, and a higher robustness against outlier measurement information. Using the Mean MSE metric and the SKF as our baseline, we found that our method consistently achieved higher estimation accuracy than strong baselines on both the bouncing ball and SLIP systems. Parallelized simulations enabled efficient large-scale experimentation, and incorporating reference-trajectory extensions enhanced estimation performance near mode transitions, and avoided state dimensional mismatches from varying state representations in different modes. Two promising directions for further work are identified. The first is the application of multi-threading for independent trajectory rollouts, along with adaptive resampling. The second is experimentation and application to hardware, perhaps in combination with learning-based methods of system identification \cite{LegRobo}.

% or


% \input{final paper/contributions}
\section*{Acknowledgment}
This material is based upon work supported by the National Science Foundation Graduate Research Fellowship Program under Grant No. DGE-2039655. Any opinions, findings, and conclusions or recommendations expressed in this material are those of the author(s) and do not necessarily reflect the views of the National Science Foundation.

\bibliography{base} % Reference your base.bib file (no extension)

\end{document}